%% file: acl_latex.tex
\pdfoutput=1

\documentclass[11pt]{article}

\usepackage[]{acl}

\usepackage{times}
\usepackage{latexsym}

\usepackage[T1]{fontenc}

\usepackage[utf8]{inputenc}

\usepackage{microtype}
\usepackage{makecell}
\usepackage{inconsolata}
\usepackage{microtype}
\usepackage{times}
\usepackage{pifont}
\usepackage{url}
\usepackage{latexsym}
\usepackage{microtype}
\usepackage{graphicx}
\usepackage{subfigure}
\usepackage{booktabs}
\usepackage{amsmath}
\usepackage{amsthm}
\usepackage{enumitem}
\usepackage{color,amsfonts}
\usepackage{multirow,array}
\usepackage{tikz}
\usepackage{pgfplots}
\usepackage{algpseudocode}
\usepackage{amssymb,mathtools}
\usepackage[linesnumbered,ruled,vlined]{algorithm2e}
\usepackage{cleveref}
\usepackage{amssymb}
\usepackage{nomencl}
\usepackage{etoolbox}
\usepackage{tcolorbox}
\usepackage{caption}
\theoremstyle{definition}
\usepackage[normalem]{ulem} 

\makeatletter
\patchcmd{\@algocf@start}
  {-1.5em}
  {0pt}
  {}{}
\makeatother

\usepackage{ulem}
\usepackage{lipsum}
\usepackage{enumitem}

\usepackage{amsthm,amsmath,amsfonts,bm,xspace}
\usepackage{color}

\theoremstyle{definition}
\newtheorem{example}{Example}

\title{Direct Evaluation of Chain-of-Thought in Multi-hop Reasoning \\ with Knowledge Graphs }

\author{Minh-Vuong Nguyen\Thanks{ Equal contribution } 
 $^{\heartsuit}$ \enskip Linhao Luo$^{*\diamondsuit}$ \enskip Fatemeh Shiri$^\diamondsuit$ \enskip Dinh Phung$^\diamondsuit,^\heartsuit$ \\ \textbf{Yuan-Fang Li$^\diamondsuit$ \enskip Thuy-Trang Vu$^\diamondsuit$ \enskip Gholamreza Haffari$^\diamondsuit$} \\
  $^\diamondsuit$Department of Data Science \& AI, Monash University \\
  $^\heartsuit$VinAI Research, Vietnam\\
   \texttt{\{trang.vu1,first.last\}@monash.edu}
}

\begin{document}
{\makeatletter\acl@finalcopytrue
  \maketitle
}
\begin{abstract}
\input{sections/0-abstract}
\end{abstract}

\input{sections/1-intro}

\input{sections/2-background}

\input{sections/3-method}

\input{sections/4-exp}
\input{sections/5-analysis}
\input{sections/7-related-work}

\input{sections/8-conclusion}
\input{sections/limitation}

\section*{Acknowledgments}
The authors are grateful to the anonymous reviewers for their helpful comments. The computational resources of this work are supported by the Multi-modal Australian ScienceS Imaging and Visualisation  Environment (MASSIVE)\footnote{\url{www.massive.org.au}}. 
This material is based on research partially sponsored by the DARPA Assured Neuro Symbolic Learning and Reasoning (ANSR) program under award number FA8750-23-2-1016 and the DARPA Knowledge Management at Scale and Speed (KMASS) program under award number HR00112220047.
Minh-Vuong Nguyen was supported by VinAI Research.

\bibliography{certreasoning}

\clearpage

\appendix
\input{sections/appendix}

\end{document}

%% file: sections/0-abstract.tex
Large language models (LLMs) have demonstrated strong reasoning abilities when prompted to generate chain-of-thought (CoT) explanations alongside answers. However, previous research on evaluating LLMs has solely focused on answer accuracy, neglecting the correctness of the generated CoT. In this paper, we delve deeper into the CoT reasoning capabilities of LLMs in multi-hop question answering by utilizing knowledge graphs (KGs). We propose a novel discriminative and generative CoT evaluation paradigm to assess LLMs' knowledge of reasoning and the accuracy of the generated CoT.
Through experiments conducted on 5 different families of LLMs across 2 multi-hop question-answering datasets, we find that LLMs possess sufficient knowledge to perform reasoning. However, there exists a significant disparity between answer accuracy and faithfulness of the CoT generated by LLMs, indicating that they often arrive at correct answers through incorrect reasoning. \footnote{Code and data are available at: \url{https://github.com/MinhVuong2000/LLMReasonCert}}

%% file: sections/1-intro.tex
\section{Introduction}
While large language models (LLMs) have shown great potential as general-purpose task solvers, they tend to be unreliable reasoners~\citep{Bang2023multitaskmultilingualmultimodal}.
Prior research suggests that LLMs demonstrate reasoning-like behaviors as the number of parameters increases~\citep{wei2022chain}. Notably, Chain-of-Thought (CoT) prompting, where LLMs are explicitly instructed to decompose questions into a sequence of logical steps before generating answers, has achieved impressive performance in various reasoning tasks~\citep{wei2022chain,kojima2022large}.
However, as LLMs function as black-box models, the mechanism behind their reasoning processes remains largely unknown.

\input{figures/error_examples}

Previous research measures the reasoning ability of LLMs by reporting their performance, e.g.\ accuracy, on the downstream tasks that require reasoning
\citep{huang-chang-2023-towards}. This evaluation strategy cannot provide a direct assessment of the reasoning steps. Hence, it remains unclear whether their strong performance is the result of true reasoning ability or simple heuristics. 
Recent studies on analyzing CoT reasoning introduce perturbations to prompts, including the injection of invalid reasoning paths, incorrect facts, or the addition of arbitrary symbols to the few-shot examples~\citep{madaan-etal-2023-makes, wang-etal-2023-towards, ye-etal-2023-assessing}. These studies show that various aspects of prompts, such as query relevance, style patterns, and the correct ordering of reasoning steps, are more important than the validity of reasoning in the in-context demonstrations. While revealing interesting insights into the reasoning process of LLMs, prompt perturbation-based methods still cannot directly evaluate the correctness of reasoning steps.
Automatically verifying CoT reasoning steps is still an open challenge due to the unstructured nature of its freeform rationales.

In this paper, we go beyond evaluating only the final answers to \textit{directly} analyzing the intermediate reasoning steps generated by CoT prompting in multi-hop question-answering (QA) tasks. To tackle the unstructured nature of CoT, we introduce a novel evaluation framework that grounds LLMs' responses in knowledge graphs (KGs) and verify whether it forms 
a faithful path to the given KGs. 
Before evaluating more open-ended generative reasoning skills, we design \textit{discriminative} tests to assess whether LLMs can identify faithful reasoning paths when perturbed with factual errors, incoherent, and misguided reasoning steps.
Our discriminative evaluation results reveal that LLMs possess certain knowledge of valid reasoning under sufficient knowledge conditions.
Building on this observation, we further propose the \textit{generative} evaluation to measure the reasoning ability of LLMs and detect fine-grained reasoning errors (see \Cref{fig:example}). In the generative evaluation, we instruct LLMs to generate CoT in a structured format, enabling us to parse their responses into a structured reasoning path and validate against KGs. Our ablation study with human experts shows that our framework achieves good accuracy in reasoning path retrieval and evaluation.

We use the proposed evaluation framework to understand the CoT reasoning process of five state-of-the-art LLMs on two complex QA tasks, which require performing multi-step reasoning to answer the questions. Our study reveals that
\begin{itemize}
    \item LLMs contain sufficient knowledge to conduct reasoning. However, they are still limited in considering the coherence of the reasoning and hallucinations during CoT generation.
    \item The correct final answer may not necessarily follow from faithful reasoning. We observe a significant gap between answer accuracy and reasoning faithfulness. It highlights the necessity of directly evaluating the reasoning steps rather than solely scoring the final answers.
    \item The performance gap between the final answer and reasoning worsens as the model size increases. As the answer accuracy also increases with the model size, it suggests that the bigger models may have the knowledge of the final answer without the need to perform reasoning.
    \item Better prompting strategies such as self-consistency or instructing LLMs with planning can further improve both the final answer and reasoning faithfulness.
\end{itemize}





%% file: figures/error_examples.tex
\begin{figure}[t]
    \centering
    \includegraphics[width=.95\columnwidth]{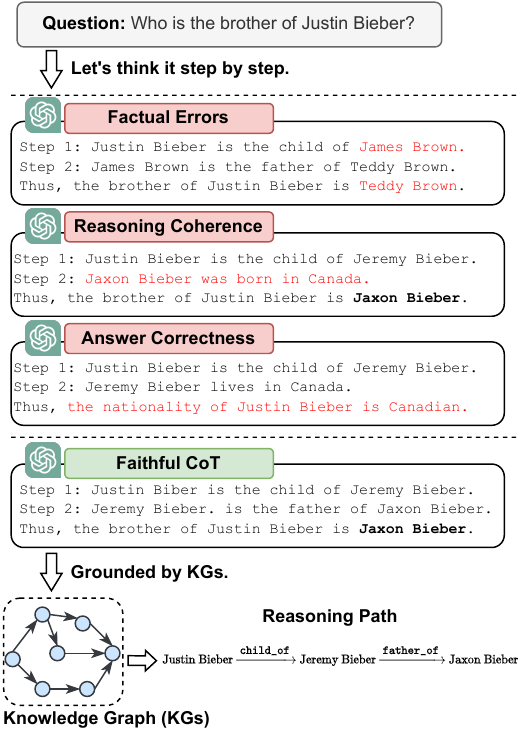}
    \caption{Examples of different reasoning errors and a faithful CoT grounded by knowledge graph.}\label{fig:example}
    \vspace{-1em}
\end{figure}

%% file: sections/2-background.tex
\section{Preliminaries}\label{sec:background}

\paragraph{Chain-of-thought (CoT) Reasoning}
\label{sec:background_cot}
Chain-of-thought (CoT) \cite{wei2022chain} is a reasoning framework that prompts LLMs to generate a step-by-step reasoning process $S=\{s_1,s_2,\ldots,s_n\}$ to a question $q$, where $s_i$ is a natural language sentence describing {a step in} the reasoning process.

\paragraph{Faithful CoT} A faithful CoT should satisfy the following properties \cite{creswell2022faithful}: (i) there are no \textit{factual} errors, (ii) the reasoning process is \textit{coherent} (i.e., the conclusion of previous step $s_{i-1}$ should be the prerequisite of the current step $s_i$), (iii) the reasoning process leads to the \textit{correct} answers. Examples of violations of these properties are shown in 
\Cref{fig:example}.

\paragraph{Knowledge Graphs (KGs)}
Knowledge graphs (KGs) are structured representations of knowledge that contain abundant facts in the form of triples $\mathcal{G}=\{(e_h,r,e_t)\mid e_h,e_t\in\mathcal{E}, r\in \mathcal{R}\}$, where $e_h$ and $e_t$ are head and tail entities, and $r$ is the relation between them; $\mathcal{E}$ and $\mathcal{R}$ are the entity and relation sets respectively. A path in KGs is a sequence of triples: $P=e_0\xrightarrow{r_1}e_1\xrightarrow{r_2}\ldots\xrightarrow{r_l}e_{l}$, {connecting the  entity $e_0$ to the  entity $e_{l}$}. 

\paragraph{Reasoning Paths}
Given a question $q$ and the answer $a$, a \textit{valid} reasoning path $P^*=e_q\xrightarrow{r_1}e_1\xrightarrow{r_2}\ldots\xrightarrow{r_l}e_{a}$ is a path that connects the topic entity $e_q$ of $q$ to the answer entity $e_a$ of $a$ in KGs. The reasoning path $P^*$ expresses a valid reasoning process for answering the question {according to the KG}.

\begin{example}
    Given a question ``Who is the brother of Justin Biber?'', we can find a valid reasoning path $P^*$ in KGs as: {
    $\text{\fontfamily{cmtt}\selectfont Justin Bieber}\xrightarrow{\texttt{child\_of}}\text{\fontfamily{cmtt}\selectfont Jeremy Bieber}\xrightarrow{\texttt{father\_of}}\text{\fontfamily{cmtt}\selectfont Jaxon Bieber}$. 
    It indicates: (i) Justin Bieber is the child of Jeremy Bieber, and (ii) Jeremy Bieber is the father of Jaxon Bieber. Thus, the brother of Justin Bieber is Jaxon Bieber.}
\end{example}




\paragraph{Faithful CoT Grounded by KGs} We verify the faithfulness of the LLMs' CoT reasoning by grounding it with KGs. By treating each reasoning step as a triple in KGs, we convert the CoT into a reasoning path. 
If a reasoning path starting from the question and ending at the answers exists in KGs, we deem the CoT of LLMs faithful. A grounded example is shown at the bottom of \Cref{fig:example}.

%% file: sections/3-method.tex
\section{Evaluating the CoT Reasoning of LLMs}\label{sec:method}
We propose  a framework to evaluate the CoT reasoning process of LLMs with the help of KGs. Specifically, we propose two evaluation modules: \textit{discriminative evaluation} and \textit{generative evaluation}. The discriminative evaluation investigates whether LLMs possess enough knowledge for conducting faithful reasoning and the generative evaluation further analyzes whether LLMs can provide faithful reasoning process during CoT generation. The overall framework is shown in \Cref{fig:framework}.

\subsection{Discriminative Evaluation}\label{sec:dis_eval}

The discriminative evaluation aims to analyze whether the LLMs possess enough knowledge to conduct faithful reasoning. i.e. whether it can recognize certain properties of faithful reasoning, including no factual error, coherence and leading to correct answers. We hypothesize that 
if the LLMs possess sufficient knowledge for faithful reasoning, they should be able to distinguish valid reasoning paths from invalid ones given the question and answer.
%
%
Following previous studies that evaluate the factual knowledge inside LLMs \cite{luo-etal-2023-systematic}, we feed both the valid and invalid reasoning paths to the LLMs and ask them to predict the validity of these paths. This allows us to assess the reasoning knowledge inside LLMs by analyzing their prediction accuracy.
We carefully design prompts to describe the task and instruct LLMs to provide the prediction. \Cref{fig:disc-template} shows an example of the zero-shot prompt template. 

\begin{figure} 
\centering
\begin{minipage}{0.95\columnwidth}
  \vspace{1mm}
  \begin{tcolorbox}[title=
  Zero-shot Discriminative Evaluation Prompt, fonttitle=\bfseries\small]
      \small
      A reasoning path is a sequence of triples that can be used to derive the answer of given question. Given this reasoning path, do you think this is a valid path to derive the answer of given question? If yes please answer "YES", otherwise please answer "NO
      \vspace{5pt}
      \\
      Question:\\
      \texttt{<Question>}
      \vspace{5pt}
      \\
      Answer:\\
      \texttt{<Answer>}
      \vspace{5pt}
      \\
      Reasoning path:\\
      \texttt{<Reasoning Path>}
  \end{tcolorbox}
  \label{box}
\end{minipage}
\caption{Discriminative Evaluation Prompt. \texttt{<Question>} indicates the question, \texttt{<Answer>} denotes the corresponding answer, and \texttt{<Reasoning Path>} denotes the input reasoning path, which is verbalized as a structured sentence}
\label{fig:disc-template}
\end{figure}


A \textit{valid} reasoning path is a sequence of triples that can
be used to derive the answer of given question. The valid reasoning paths are extracted from the ground-truth reasoning paths\footnote{The ground-truth reasoning paths are constructed from the SPARQL queries provided in the datasets. The detailed construction is shown in \Cref{appendix:subgraph_gr_construct}.} $P^*\in\mathcal{P}^*$. 
%
We generate three types of \textit{invalid} reasoning paths $P'$ by breaking specific properties of a faithful CoT: 
\begin{itemize}
    \item \textit{Factual error reasoning path:} we construct the invalid paths with factual errors by randomly corrupting entities within the valid reasoning path. This would result in some factual errors in the reasoning path, which are not valid for answering the question.
    \item \textit{Incoherent reasoning path:} we shuffle the triples of valid paths to construct an incoherent reasoning path. Even though the facts within the paths are accurate, the overall coherence of the path is compromised.
    \item \textit{Misguided reasoning path: } we randomly sample the paths starting from other questions in KGs. These paths are factually correct and coherent, but they are not related to the questions and lead to incorrect answers.
\end{itemize}




\input{figures/framework}

To thoroughly assess the reasoning abilities of LLMs, in addition to the zero-shot prompt, we have also developed few-shot, zero-shot CoT, and few-shot CoT prompts. The details of these prompts are shown in \Cref{appendix:dis_prompt}.

\paragraph{Findings} The results of the discriminative assessment are shown in \S\ref{sec:dis_results}. From the results, we can conclude that LLMs possess enough knowledge to identify factual errors as well as reasoning path relatedness, but have limitations in considering the coherence of reasoning paths and CoT generation. Therefore, we propose the generative evaluation to further assess the faithfulness of CoT reasoning in LLMs' generation.

\subsection{Generative Evaluation}\label{sec:gen_eval}
The generative evaluation aims to assess the faithfulness of the CoT reasoning process \textit{generated} by LLMs. 
Our main idea is to \textit{ground} LLMs' CoT into KG and verify whether it forms a valid path. 
%
%
To address the challenge of evaluating unstructured CoT, we carefully design a prompting strategy to instruct LLMs to output the CoT in a structured format. This enables us to parse LLM's response into a structured reasoning path, which can then be validated against KG. The example prompts and structured CoT output are provided in \Cref{appendix:gen-prompt}.

Specifically, given a generated CoT response $S$ of question $q$, we first construct a reasoning path $\hat{P}$ by retrieving triples from the KGs. Then, we evaluate the validity of the reasoning path by checking whether the path coherently connects the question and answer entities in the KGs. The details of these steps are explained in the following subsections. 

\subsubsection{Reasoning Path Construction}\label{sec:path_construct}
Given a CoT response $S=\{s_1,s_2,\ldots,s_n\}$, we first retrieve a triple\footnote{{We noticed in almost all cases in our experiments, a CoT step corresponds to one KG triplet. The extension to multiple triplets per CoT step is left for future work.} 
} $T=(e_h,r,e_t)$ for each step $s_i$ in the CoT response. 
The retrieved triple is the structural representation of each reasoning step, which can be used to construct the reasoning path for evaluation.

Previous works usually retrieve triples by identifying the entities and relations mentioned in the sentence and linking them to the KGs \cite{ijcai2021p611,
wang-etal-2021-retrieval}. However, this process is not scalable to
KGs. Inspired by the recent fact retrieval method \cite{baek2023direct}, we represent the reasoning step $s$ and triples in a unified embedding space and retrieve the triple $T$ based on their embedding similarity.  

For all the triples in a KG $\mathcal{G}$, we verbalize each triplet into a sentence by concatenating the entities and relation $x="e_h~r~e_t."$. Then, we use the Sentence-BERT model \citep{reimers-2019-sentence-bert} to obtain its embedding $h_T=E(x)$. These embeddings are constructed in advance and saved in a vector database 
for efficient retrieval. Similarly, the embedding of a given reasoning step $s$ is computed as $h_{s} = E(s)$. Then we retrieve the top-$K$ triples from KG by calculating the embedding similarity between $h_{s}$ and $h_T$ as:
\begin{equation}
  \centering
  \tau_i = f(h_{s}, h_{T_i}), \quad T_i=(e_h,r,e_t)\in \mathcal{G},
\end{equation}
where $\tau_i$ denotes the similarity score of triple $T_i$, and $f(\cdot,\cdot)$ is a non-parametric scoring function that measures the similarity between two embeddings. We adopt cosine similarity as the scoring function.

The embedding-based retrieval method may lead to the omission of entities mentioned in the reasoning step. To solve this problem, we also take into account the presence of head and tail entities in the reasoning step in the scoring function. The final score for each retrieved triple is calculated as,
\begin{equation}
  \centering
  \tilde{\tau_i} = \frac{\tau_i + \epsilon_h + \epsilon_t}{3},
\end{equation}
where $\epsilon_h$ and $\epsilon_t$ represent the fuzzy-match ratio
of head and tail entities in the reasoning step, which are range from 0 to 1, where 0 denotes no existence, 1 denotes a complete match. The overall retrieval process is presented in the \Cref{algo:retrieval} in the \Cref{app:gen_eval}. 

Thus, we could obtain a set of triples $\mathcal{T}=\{T_1,T_2,\ldots,T_n\}$ for the CoT response $S$. Then, we construct the reasoning path $\hat{P}$ by connecting the triples in $\mathcal{T}$.

\subsubsection{Reasoning Path Evaluation}\label{sec:reasoning_eval}
    By evaluating the validity of constructed reasoning paths, we can assess the faithfulness of the CoT reasoning process generated by LLMs. Specifically, we evaluate the validity of the constructed path $\hat{P}$ from three aspects: 
    \begin{itemize}
        \item \textit{Factual correctness:}  
        $\hat{P}$ contains factual error if the similarity score $\tilde{\tau_i}$ of any retrieved triples are below a factual threshold $\sigma$.
        \item \textit{Coherence:} given a factually correct path, it is incoherent if there exists a step where its premise is not the conclusion of the previous step.
        \item \textit{Final answer correctness: } given a factually correct and coherent path, whether the final answer is correct, i.e. matched with ground-truths.
    \end{itemize}
\paragraph{Validity of Reasoning Path} The prerequisite and conclusion at each reasoning step are considered head and tail entities, respectively. If the reasoning path $\hat{P}$ can connect the question and answer entities in the KG, we can conclude that it is a valid path. The detailed algorithm is shown in the \Cref{algo:eval} in the \Cref{app:gen_eval}.

\paragraph{Fine-grained Assessment} 
In addition to the binary evaluation, we also report the minimum edit distance between the constructed reasoning path $\hat{P}$ and the ground-truth path $P^*$. This serves as a fine-grained assessment of CoT reasoning capability.
We adopt a widely used sequence alignment algorithm - Needleman Wunsch algorithm \citep{needleman1970general} to
obtain continuous alignment scores (i.e., edit distance), which indicate how close the constructed reasoning path is to the ground-truth reasoning paths. If multiple ground-truth paths exist, we report the score against one with the highest match rate. The detailed algorithm is shown in the \Cref{algo:eval_groundtruth} in the \Cref{app:gen_eval}.


%% file: figures/framework.tex
\begin{figure*}[t]
    \centering
    \includegraphics[width=.9\textwidth]{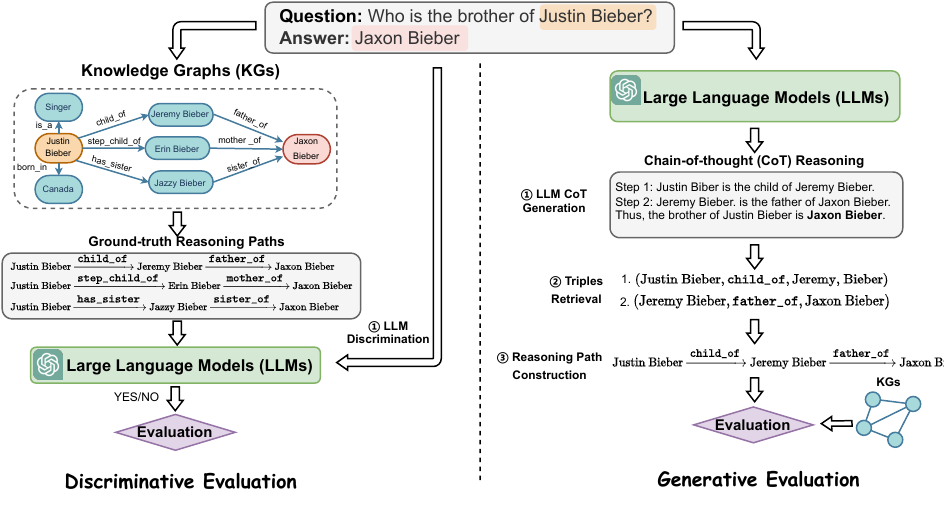}
    \caption{The overall framework of evaluating the CoT reasoning of LLMs, which contains two evaluation modules: discriminative evaluation and generative evaluation. The orange and red rectangles denote the entities mentioned in the question and answer, respectively.}
    \label{fig:framework}
\vspace{-1em}
\end{figure*}

%% file: sections/4-exp.tex
\section{Experiment Settings}
We use the proposed evaluations to understand the CoT reasoning process of the state-of-the-art LLMs on complex question-answering (QA) tasks which requires performing multi-step reasoning to answer the questions. Through analysis, we seek to answer the following research questions (\textbf{RQs})
\begin{itemize}
    \item \textbf{RQ1}: 
    \textit{Do LLMs have the knowledge of faithful reasoning?}
    We leverage the discriminative evaluation to test whether LLMs can identify valid reasoning paths. This evaluation focuses on assessing LLMs' knowledge about the properties of
    faithful reasoning described in~\Cref{sec:background}.
    
    \item \textbf{RQ2}: \textit{Can LLMs express such knowledge to generate faithful reasoning?}
    Utilizing our generative evaluation framework, we assess the capacity of LLMs to produce coherent and correct reasoning.  We also investigate various factors, such as model size and prompting strategies, to understand their impact on  reasoning capability.
\end{itemize}

\input{table/dataset_stat}


\paragraph{Dataset}
We 
conduct experiments on two QA datasets: Complex WebQuestions (CWQ) \cite{Talmor_2018} and GrailQA \cite{gu2021beyond} which contain up to 4-hop questions.
%
To evaluate multi-step reasoning capability, we filter out single-hop questions in the test set. \Cref{tab:stat_data} shows the statistics of the filtered test set. 
%
The generated reasoning paths are validated against Freebase \cite{bollacker2008freebase} - an open knowledge graph containing
around 88M entities, 20K relations, and 126M triples. More details can be found at \Cref{appendix:dataset}.




\paragraph{Large Language Models} 

We evaluate the reasoning capability of several LLMs with instruction-following capability at different sizes, including \texttt{Mistral} (7B) \cite{jiang2023mistral}, \texttt{Qwen} (7B, 14B) \cite{qwen}, \texttt{Vicuna} (33B) \cite{vicuna2023}, \texttt{LLaMA2-Chat} (70B) \cite{touvron2023llama} and \texttt{ChatGPT} (175B)\footnote{Previous works mentioned ChatGPT having 175B parameters~\citep{meyer2023chatgpt}. However, OpenAI still doesn’t give any official news about ChatGPT’s model size.}\cite{openai2023gpt4}.
The details of model versions are available in~\Cref{appendix:llm}.
We set temperature as 0.7 and top $p$ as 0.9 for generation in all models.

\paragraph{Prompting Strategies} We experiment with multiple CoT prompting strategies, including
\begin{itemize}
    \item \textbf{Few-shot CoT} Five examples with structured CoT followed by the answer are added to the prompt (\Cref{fig:fewshot_cot_only_prompt} in ~\Cref{appendix:gen-prompt}).
    \item \textbf{Few-shot CoT with planning (CoT-Plan)} We also explore the ability of LLMs to plan and decompose the relations required to reach the answer before verbalizing the CoT reasoning. In particular, we add the ground-truth \textit{plan} \cite{luo_rog} (i.e., a relation path pointing to the answers) into each example. An example prompt is given in \Cref{fig:fewshot_cot_hint_prompt} in ~\Cref{appendix:gen-prompt}.
    \item \textbf{Few-shot CoT with self-consistency (CoT-SC)} Beyond the conventional CoT prompting, we also experiment with Self-Consistency~\citep{Wang2022Selfconsistencyimproves}, a more sophisticated method designed to mitigate the inconsistencies in CoT reasoning by aggregating the final answer through majority votes.  In our evaluation, we sample four outputs, 
    and report the maximum performance across all the outputs.
\end{itemize}



\paragraph{Evaluation Framework Implementation} 
Given a question from the benchmark, 
in discriminative evaluation, we construct the invalid paths by randomly 
perturbing the ground-truth paths extracted from SPARQL~\citep{Kumar2019DifficultyControllableMQ}. The implementation detail 
is described in Appendix \ref{appendix:subgraph_gr_construct}. In generative mode, we use FAISS \citep{johnson2019billion} as the vector database, Sentence-BERT \citep{reimers-2019-sentence-bert}  as the employed embedding model and partial ratio fuzzy matching\footnote{\url{https://github.com/seatgeek/thefuzz}} 
as the entity scoring function. We retrieve top-$10$ triples and set the factual threshold $\sigma$ of $0.7$.

\paragraph{Evaluation Metrics}
For discriminative evaluation, we report the accuracy of detecting valid reasoning paths from invalid ones. For generative evaluation, we report CoT reasoning performance of LLMs with the following metrics: (i) final answer accuracy, (ii) faithful reasoning score, and (iii) minimum edit distance between the generated and ground truth paths. 
As different LLMs vary in instruction-following capabilities and guardrail implementations, we may encounter responses with unstructured format or abstained answers \cite{luo-etal-2023-systematic}. Therefore, we classify LLMs' responses into four groups: abstained (A), unstructured (U), faithful reasoning (FR), and unfaithful reasoning (UR). We use the F1 score to measure the faithfulness of CoT reasoning where precision and recall are calculated as $P = \frac{\textrm{FR}}{\textrm{FR} + \textrm{UR}}$ and $R = \frac{\textrm{FR}}{\textrm{FR} + \textrm{UR}  + \textrm{A}  + \textrm{U}}$.
Detailed implementations are described in \Cref{appendix:eval-metric}.  Results of precision and recall are presented in \Cref{sec:precision,sec:recall}. 

\section{Main Results}

\input{table/discriminative-results-main}

\subsection{Discriminate Evaluation}\label{sec:dis_results}
\input{table/generative_cert_main}
\paragraph{Finding 1: LLMs possess knowledge of valid reasoning}
The overall discriminative evaluation results are shown in \Cref{all_dis_result}. Based on the results, it is evident that all LLMs achieve a high level of accuracy in distinguishing valid reasoning paths. This indicates that LLMs, which are pre-trained on large-scale corpora, already possess certain knowledge to perform reasoning tasks effectively. However, when using few-shot prompts, there is a noticeable decrease in performance for \texttt{Mistral} and \texttt{LLaMA2}. This could be attributed to the sensitivity of these particular LLMs towards the provided few-shot examples. The detailed results of each perturbation type are illustrated in \Cref{sec:detailed_dis}, where the accuracy of incoherent paths is lower than other types. We speculate that LLMs cannot capture structural information in the context \cite{guo2023gpt4graph}. Moreover, the few-shot CoT fails to improve the accuracy in identifying valid paths. We speculate that LLMs are prone to hallucination during CoT generation, resulting in incorrect predictions. 
We can conclude that despite having enough reasoning knowledge, LLMs still face limitations in conducting faithful reasoning during CoT generation.


\subsection{Generative Evaluation}

Table \ref{tab:generative-cert} shows the performance of LLMs in generative evaluation mode. 
Overall, \texttt{ChatGPT} demonstrate superior performance in terms of both final answer accuracy and faithfulness of the reasoning. 
Surprisingly, \texttt{Mistral 7B}, despite being the smallest model, exhibits competitive performance comparable to larger models within the <50B range. Furthermore, enhancing prompting strategies with planning (CoT-Plan) and self-consistency (CoT-SC) results in substantial improvements across all LLMs, especially for smaller models.


\paragraph{Finding 2: The correct final answer may not necessarily result from faithful reasoning}
As shown in ~\Cref{tab:generative-cert}, there is a notable discrepancy between the accuracy of the final answer and the reasoning process. The average gap is 15.76\% for CWQ, and 16.44\% for GrailQA. 
While advanced prompting may improve answer and reasoning accuracy, this performance gap 
mostly stays consistent. 
Interestingly, \texttt{Vicuna} achieves reasonable answer accuracy but has the lowest reasoning performance of all the models, suggesting its reasoning ability is inferior, even when compared to small models like \texttt{Mistral} and \texttt{Qwen-7B}. 
This finding highlights the inadequacy of relying on final answer accuracy as a proxy to gauge reasoning ability.


\input{figures/type_of_error}

\paragraph{Finding 3: The reasoning gap worsens as the model size increases} It can be seen that the reasoning performance increases gradually with model size, proving the reasoning ability of bigger models. However, the gap between answer and reasoning performance also gradually increases with model size and the correctness of the answer. While \texttt{LLaMA2-70B} and \texttt{ChatGPT} rank first in performance,  their gaps are also the highest. Meanwhile, the smallest-size models, including \texttt{Mistral-7B} and \texttt{Qwen-7B}, hold the lowest gap on CWQ and GrailQA, respectively. 
We speculate that larger LLMs may grasp the question context better or have more knowledge to provide the correct answer directly without 
performing reasoning.

\paragraph{Finding 4: Better prompting strategy can improve both the answer and reasoning accuracy without worsening the reasoning gap} 
The use of enhanced prompting strategies such as CoT-Plan and CoT-SC lead to improvement of both the answer and reasoning accuracy across most LLMs.
However, the gap between them remains consistent, regardless of prompt strategy.

%% file: table/dataset_stat.tex
\begin{table}
\centering
\resizebox{.7\columnwidth}{!}{%
\begin{tabular}{@{}c|rrr@{}}
\hline
Dataset & \textbf{\#Test} & \textbf{\#2hop} & \textbf{\#$\geq$3hop}\\
\hline
CWQ & 1421 & 1386 & 35 \\
GrailQA & 1813 &  1528 & 285 \\
\hline
\end{tabular}
}
\caption{Statistic of datasets.}
\label{tab:stat_data}
\vspace{-1em}
\end{table}

%% file: table/discriminative-results-main.tex
\begin{table}[]
\resizebox{\columnwidth}{!}{%
\begin{tabular}{@{}l|r|cccc@{}}
\toprule
\multicolumn{1}{c}{LLMs}        & \multicolumn{1}{|c}{Size} & \multicolumn{1}{|c}{Zero-shot} & \multicolumn{1}{c}{Zero-shot CoT} & \multicolumn{1}{c}{Few-shot} & \multicolumn{1}{c}{Few-shot CoT} \\ \midrule
 \texttt{Mistral}    & 7B   & 87.59   & \underline{89.88}       & 56.91  & 69.98      \\
\texttt{Qwen}        & 7B   & 74.76   & 76.13       & \underline{79.64}  & 73.23      \\
\texttt{Qwen}        & 14B  & 88.59   & \underline{88.86}       & \textbf{88.81}  & 75.87      \\
\texttt{Vicuna-1.5}  & 33B  & \textbf{92.79}   & \textbf{\underline{92.88}}       & 84.91  & 67.05      \\
\texttt{LLaMA2-Chat} & 70B  & 77.96   & \underline{80.71}       & 56.99  & 47.76      \\
\texttt{ChatGPT}     & 175B & 89.86   & \underline{90.17}       & 87.09  & \textbf{80.15}      \\ \bottomrule
\end{tabular}%
}\caption{Discriminative evaluation results of different LLMs on CWQ. We use binary accuracy as the metric. The best results of each column and row are highlighted in \textbf{bold} and \underline{underlined}.}\label{all_dis_result}
\vspace{-1em}
\end{table}

%% file: table/generative_cert_main.tex
\begin{table*}[t!]
\centering
\resizebox{.9\textwidth}{!}{%
\begin{tabular}{@{}l|r|rrrr|rrrr@{}}  
\toprule
\multirow{2}{*}{LLMs} & \multirow{2}{*}{Size} & \multicolumn{4}{c}{\textbf{CWQ}} & \multicolumn{4}{c}{\textbf{GrailQA}}\\
 &  & \multicolumn{1}{c}{Answer$\uparrow$} & Reasoning$\uparrow$ & \multirow{1}{*}{Gap$\downarrow$} & \multirow{1}{*}{Edit Dist.$\downarrow$} & 
\multicolumn{1}{c}{Answer$\uparrow$} & \multicolumn{1}{c}{Reasoning$\uparrow$} & \multirow{1}{*}{Gap$\downarrow$} & \multirow{1}{*}{Edit Dist.$\downarrow$}\\
\midrule 
& & \multicolumn{8}{c}{Fewshot CoT} \\  
\midrule 
 \texttt{Mistral} & 7B & \multicolumn{1}{c}{36.45} & 25.18 & \multicolumn{1}{c}{\textbf{11.27}} & 69.86 & 16.35 & 2.12 & 14.23 & 94.03\\ 
  \texttt{Qwen} & 7B & \multicolumn{1}{c}{32.52} & 19.38 & \multicolumn{1}{c}{13.14} & 76.78 & 13.35 & 1.63 & \textbf{11.72} & 94.69\\
  \texttt{Qwen}  & 14B & \multicolumn{1}{c}{40.39} & 27.38 & \multicolumn{1}{c}{13.01} & 74.49 & 18.83 & 2.13 & 16.70 & 92.90\\
  \texttt{Vicuna} & 33B & \multicolumn{1}{c}{44.50} & 15.92 & \multicolumn{1}{c}{28.58} & 74.60 & 18.26 & 0.95 & 17.31 & 95.39\\ 
 \texttt{LLaMA2} & 70B & \multicolumn{1}{c}{49.80} & 33.98 & \multicolumn{1}{c}{15.82} & 62.23 & 22.05 & 2.88 & 19.17 & 92.58\\ 
 \texttt{ChatGPT} & 175B & \multicolumn{1}{c}{\textbf{49.85}} & \textbf{37.13} & \multicolumn{1}{c}{12.72} & \textbf{57.94} & \textbf{23.69} & \textbf{4.17 }& 19.52 & \textbf{90.13}\\  \midrule

& & \multicolumn{6}{c}{Fewshot CoT - Plan} \\ 
\midrule
  \texttt{Mistral} & 7B & \multicolumn{1}{c}{37.14\fontsize{6}{1}\selectfont \textcolor{black}{+0.69}} & 25.69\fontsize{6}{1}\selectfont \textcolor{black}{+0.51} & \multicolumn{1}{c}{\textbf{11.45}} & 70.01 & 17.30\fontsize{6}{1}\selectfont \textcolor{black}{+0.95} & 3.36\fontsize{6}{1}\selectfont \textcolor{black}{+1.24} & 13.94 & 94.46\\
  \texttt{Qwen}  & 7B & \multicolumn{1}{c}{35.35\fontsize{6}{1}\selectfont \textcolor{black}{+2.91}} & 21.57\fontsize{6}{1}\selectfont \textcolor{black}{+2.19} & \multicolumn{1}{c}{13.86} & 74.74 & 13.74\fontsize{6}{1}\selectfont \textcolor{black}{+0.39} & 2.06\fontsize{6}{1}\selectfont \textcolor{black}{+0.43} & \textbf{11.68} & 94.61\\
  \texttt{Qwen}  & 14B & \multicolumn{1}{c}{40.86\fontsize{6}{1}\selectfont \textcolor{black}{+0.47}} & 27.97\fontsize{6}{1}\selectfont \textcolor{black}{+0.59} & \multicolumn{1}{c}{12.02} & 73.68 & 19.00\fontsize{6}{1}\selectfont \textcolor{black}{+0.17} & 2.48\fontsize{6}{1}\selectfont \textcolor{black}{+0.35} & 15.43 & 92.58\\
  \texttt{Vicuna} & 33B & \multicolumn{1}{c}{48.80\fontsize{6}{1}\selectfont \textcolor{black}{+4.30}} & 
 20.24{\fontsize{6}{1}\selectfont \textcolor{black}{+4.32}}
 & \multicolumn{1}{c}{28.56} & 63.93 & 
 20.84{\fontsize{6}{1}\selectfont \textcolor{black}{+2.58}}
 & 2.09\fontsize{6}{1}\selectfont \textcolor{black}{+1.14} & 18.75 & 92.12\\
 \texttt{LLaMA2}  & 70B & 
\multicolumn{1}{c}{50.26{\fontsize{6}{1}\selectfont \textcolor{black}{+0.46}}}
& 
37.08{\fontsize{6}{1}\selectfont \textcolor{black}{+3.10}}
 & \multicolumn{1}{c}{13.18} & 57.81 &
22.35{\fontsize{6}{1}\selectfont \textcolor{black}{+0.30}}
& 3.29\fontsize{6}{1}\selectfont \textcolor{black}{+0.41}
& 19.06 & 89.61\\ 
 \texttt{ChatGPT} & 175B & \multicolumn{1}{c}{\textbf{51.74{\fontsize{6}{1}\selectfont \textcolor{black}{+1.89}}}} & \textbf{38.60{\fontsize{6}{1}\selectfont \textcolor{black}{+1.47}}} & \multicolumn{1}{c}{13.14} & \textbf{56.61} & \textbf{24.21{\fontsize{6}{1}\selectfont \textcolor{black}{+0.52}}} & \textbf{4.32{\fontsize{6}{1}\selectfont \textcolor{black}{+0.15}}} & 19.11 & \textbf{89.84}\\ 
 \midrule	
 
& & \multicolumn{6}{c}{Fewshot CoT - SC} \\ 
\midrule
  \texttt{Mistral}  & 7B & \multicolumn{1}{c}{40.86\fontsize{6}{1}\selectfont \textcolor{black}{+4.41}} & 30.38\fontsize{6}{1}\selectfont\textcolor{black}{+5.20} & \multicolumn{1}{c}{\textbf{10.48}} & 65.21 & 16.70{\fontsize{6}{1}\selectfont \textcolor{black}{+0.35}} & 2.60{\fontsize{6}{1}\selectfont \textcolor{black}{+0.48}} & 14.10 & 94.10\\
  \texttt{Qwen}  & 7B & \multicolumn{1}{c}{34.75\fontsize{6}{1}\selectfont\textcolor{black}{+6.08}} & 23.21\fontsize{6}{1}\selectfont \textcolor{black}{+3.83} & \multicolumn{1}{c}{15.39} & 74.24 & 14.00{\fontsize{6}{1}\selectfont \textcolor{black}{+0.65}} & 2.32{\fontsize{6}{1}\selectfont \textcolor{black}{+0.69}} & \textbf{11.68} & 94.35\\
  \texttt{Qwen}  & 14B & \multicolumn{1}{c}{41.01\fontsize{6}{1}\selectfont \textcolor{black}{+0.62}} & 29.26\fontsize{6}{1}\selectfont \textcolor{black}{+1.88} & \multicolumn{1}{c}{11.75} & 73.21 & 21.00{\fontsize{6}{1}\selectfont \textcolor{black}{+2.17}} & 3.24{\fontsize{6}{1}\selectfont \textcolor{black}{+1.11}} & 17.76 & 92.50\\
  \texttt{Vicuna} & 33B & \multicolumn{1}{c}{45.43\fontsize{6}{1}\selectfont \textcolor{black}{+2.18}} & 21.32\fontsize{6}{1}\selectfont \textcolor{black}{+5.40} & \multicolumn{1}{c}{25.36} & 66.17 & 18.92{\fontsize{6}{1}\selectfont \textcolor{black}{+0.66}} & 1.88{\fontsize{6}{1}\selectfont \textcolor{black}{+0.93}} & 17.04 & 94.23\\
 \texttt{LLaMA2} & 70B & \multicolumn{1}{c}{50.42\fontsize{6}{1}\selectfont \textcolor{black}{+0.62}} & 37.00\fontsize{6}{1}\selectfont \textcolor{black}{+3.02} & \multicolumn{1}{c}{13.42} & 58.55	& 22.35{\fontsize{6}{1}\selectfont \textcolor{black}{+0.30}} & 3.29{\fontsize{6}{1}\selectfont \textcolor{black}{+0.41}} & 19.06 & 91.50\\ 
 \texttt{ChatGPT} & 175B & \multicolumn{1}{c}{\textbf{51.74\fontsize{6}{1}\selectfont \textcolor{black}{+1.89}}} & \textbf{40.73{\fontsize{6}{1}\selectfont\textcolor{black}{+3.60}}} & \multicolumn{1}{c}{11.01} & \textbf{52.57} & \textbf{24.97{\fontsize{6}{1}\selectfont \textcolor{black}{+1.28}}} & \textbf{4.86{\fontsize{6}{1}\selectfont \textcolor{black}{+0.69}}} & 20.11 & \textbf{89.22}\\ 
 \bottomrule
\end{tabular}
}
\caption{
Generative evaluation performance of different LLMs on CWQ and GrailQA datasets. F1-scores of the final answer and reasoning accuracy are reported in \texttt{Answer} \texttt{Reasoning} respectively. The \texttt{Gap} column denotes the differences between \texttt{Answer} and \texttt{Reasoning}. The \texttt{Edit Dist.} denotes the edit distance metric described in \Cref{appendix:eval-metric}. \texttt{+x.xx} denotes the improvement in comparison to few-shot CoT.}
\label{tab:generative-cert}
\vspace{-1em}
\end{table*}

%% file: figures/type_of_error.tex
\begin{figure*}
   \centering
   \resizebox{0.9\textwidth}{!}{%
    \begin{tabular}{cc}
        \includegraphics[width=8cm]{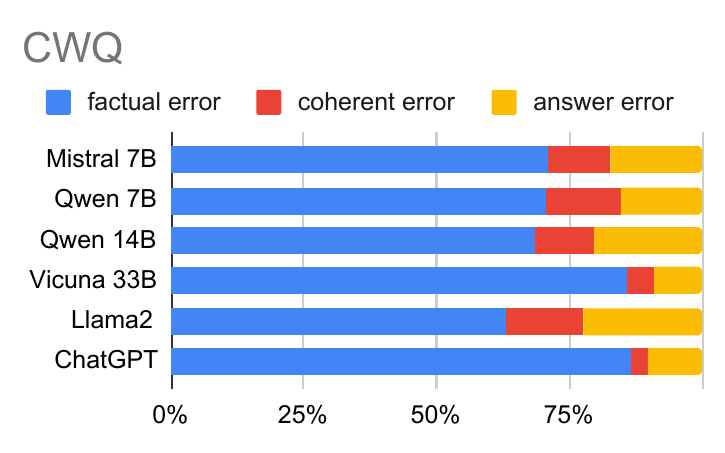}
        &
        \includegraphics[width=8cm]{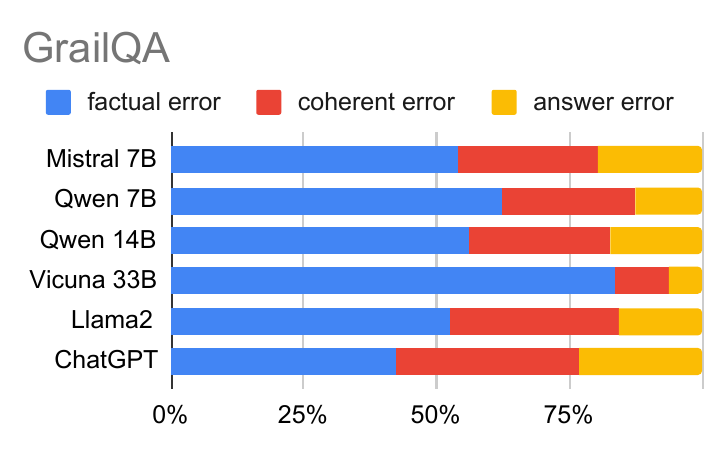}
    \end{tabular}
    } 
    \caption{The breakdown of reasoning error types in CWQ and GrailQA.}
    \label{fig:err_typs} 
\vspace{-1em}
\end{figure*}

%% file: sections/5-analysis.tex
\subsection{Analysis}
\paragraph{Reasoning Errors} 
We present a detailed breakdown of the reasoning errors in \Cref{fig:err_typs}. 
The results reveal that factual errors account for the majority of errors, indicating that LLMs tend to generate incorrect information during reasoning.
As GrailQA is a more complex dataset, LLMs have a higher percentage of coherence errors on GrailQA than CWQ.
Interestingly, even when the generated CoT paths are free from factual and coherent errors, LLMs may fail to produce correct answers, evidenced by a substantial amount of answer errors. Error case examples are shown in \Cref{appendix:case-study}.

\paragraph{Ablation Study}
To ensure the effectiveness of our generative evaluation framework, we randomly select 100 CoT responses generated by \texttt{ChatGPT} in CWQ dataset and asked two human experts to evaluate the constructed reasoning path. The detail of the human evaluation study is described in the appendix \ref{sec:human_eval}.
The results presented in \Cref{tab:eval_framework} demonstrate that our method can accurately detect both faithful and unfaithful reasoning paths. This further confirms the efficacy of our approach in evaluating CoT reasoning.

\input{table/eval_framework_annotation}

\paragraph{Parsing Error}
While we carefully design prompts to instruct LLMs to generate a structured CoT, there are still corner cases where LLMs generate unstructured and abstention responses due to their unpredicted behaviors. As reported in \Cref{sec:precision}, the unstructured and abstention rates are less than 20\% in CWQ dataset and can be mitigated with CoT-Plan and CoT-SC.

%% file: table/eval_framework_annotation.tex
\begin{table}
\centering
\resizebox{0.9\columnwidth}{!}{%
\begin{tabular}{@{}l|ccc@{}}  
\toprule
 Reasoning Types & Precision & Recall & F1\\
\midrule 
 Faithful Reasoning  & 95.42 & 83.88 & 89.28\\
 Unfaithful Reasoning & 86.12 & 96.01 & 90.80\\
 \bottomrule
\end{tabular}
}
\caption{Precision, Recall, and F1 score of the framework on human annotated datasets.}
\label{tab:eval_framework}
\vspace{-1em}
\end{table}

%% file: sections/7-related-work.tex
\section{Related Work}
\paragraph{Reasoning with LLMs}
While LLMs have proven to offer a variety of reasoning abilities, they still tend to hallucinate facts, making them unreliable and imperfect~\citep{qiao2022reasoning}. Several studies have concentrated on improving their reasoning capacity through prompting~\citep{Wang2022Selfconsistencyimproves, Ye2022UnreliabilityExplanationsFew, Wiegreffe2022ReframingHumanAI}. CoT~\citep{wei2022chain} is a prompting approach that has demonstrated notable improvements in reasoning performance. A significant enhancement compared to CoT, self-consistency~\citep{Wang2022Selfconsistencyimproves}, is a scheme where multiple CoTs are generated and the most consistent self-generated answer is selected. Recently, self-consistency was extended with Tree of Thoughts (ToT)~\citep{yao2023tree}, which models the reasoning process with a tree. ToT allows LLMs to interactively backtrack and explore alternate chains of reasoning, avoiding getting stuck on a single line of incorrect reasoning. \citet{Ye2022UnreliabilityExplanationsFew} mitigate the effect of unreliable rationales by calibrating the prediction probability based on the factuality of CoT. \citet{Wiegreffe2022ReframingHumanAI} train a Seq2Seq model to filter out unacceptable rationale. \citet{liu2021generated} utilize GPT-3~\citep{brown2020language} with few-shot prompting to generate knowledge and prompts the downstream language models.

\paragraph{Reasoning Evaluation}


Evaluation of the reasoning ability of LLMs has been undertaken for two main purposes: to enhance the reasoning ability of LLMs \citep{lyu2023faithful,li2023making,tyen2023llms,chen2023felm} and to quantify the reasoning ability of LLMs~\citep{wang2023boosting,Atanasova2023FaithfulnessTestsNatural}. For instance, \citet{huang-chang-2023-towards} gauges the reasoning ability of LLMs by assessing their performance on reasoning benchmarks such as GSM8K and BIG-bench for downstream tasks. However, this evaluation strategy is unable to offer a direct assessment of the reasoning steps. \citet{tyen2023llms} release the BIG-Bench Mistake dataset, which includes logical errors in CoT reasoning steps. Using this benchmark, \citet{tyen2023llms, chen2023felm} illustrate the inability of state-of-the-art LLMs to identify mistakes and reasoning errors, even in unequivocal cases.

%% file: sections/8-conclusion.tex
\section{Conclusion}
We propose an evaluation framework to understand the CoT reasoning capability of LLMs beyond the sole assessment of final answer accuracy.
With the help of a KG and a careful prompting strategy, we can turn the unstructured CoT into a structured format for automatic evaluation.
Our framework consists of two evaluation modules: (i) a discriminative module that isolates the effects of different reasoning errors 
to verify LLMs' knowledge about reasoning, and (ii) a generative module to assess the generated CoT reasoning.
While LLMs showcase remarkable capabilities in generating correct answers, our study emphasizes the need for more nuanced evaluations of their reasoning processes. 
Addressing the gap between the final answer and reasoning accuracy remains a critical area for further exploration in enhancing the true reasoning capabilities and interpretability of LLMs.

%% file: sections/limitation.tex
\section*{Limitation}
The limitation of our work includes:
\begin{itemize}
    \item We consider a CoT step corresponding to one KG triple and a single correct answer for each question. However, LLMs may generate a sentence containing more than two relations.
    This can be tackled by returning top-K candidates from \Cref{algo:retrieval} and a dynamic program algorithm expanded from \Cref{algo:eval}.
    \item We assume the availability of completed knowledge graphs (KGs) for factual retrieval.
    We leave the incorporation of the knowledge graph completion methods to improve the comprehension of the retrieval algorithm as future works.
    \item This study mainly focuses multi-hop reasoning questions over knowledge graphs (KGs). Nevertheless, there exist intricate reasoning inquiries, such as those in mathematics or logical reasoning, which involve unstructured replies that are not easily resolvable through KGs. The establishment of verification frameworks for diverse forms of reasoning queries plays a significant role in enhancing the reliability and utility of responses generated by LLMs. This aspect is still an open and challenging problem, requiring extensive explorations within the research community.
\end{itemize}

%% file: sections/appendix.tex
\label{sec:appendix}
\input{sections/appx/appx-generative-eval}

\input{sections/appx/appx-experiment-setting}

\section{Additional Results}
\label{appendix:results}

\subsection{Detailed Results of Discriminative Evaluation}\label{sec:detailed_dis}
\input{table/detailed_discriminative_evaluation}

We illustrate the discriminative evaluation results of different types of reasoning paths in \Cref{dis_result}. For each error types, LLMs reach the best performance under factual errors and misguided reasoning path. However, the performance under incoherent paths is lower, which could due to the limits of LLMs in understanding structural context. Besides, the performance of valid reasoning path is slightly lower than invalid reasoning path. Because it requires both three properties (i.e., factual errors, incoherence, misguidance) satisfied at the same time.

\subsection{Detailed Results of Precision}\label{sec:precision}
\input{figures/prec_cot_only}
\input{figures/prec_cot_hint}
\input{figures/prec_cot_sc}

Based on the definition in \cref{eq:precision}, precision ignores the abstained and unstructured responses. The detailed precision results as well as the abstained and unstructured ratio are shown in \Cref{fig:prec_cot_only,fig:prec_hint,fig:prec_cot_sc}.

 The precision of the answer gradually improves with the size of the model. However, Mistral 7B and Qwen 14B are competitive in terms of reasoning performance compared to models with ten times more parameters. While Vicuna 33B has an intermediate model size, its performance is low. It is also worth noting that Qwen variants exhibit a high rate of abstention. This leads to abstaining from answering uncertain questions and achieving higher precision, but at the cost of reduced recall for correct answers and reasoning.


\subsection{Detailed Results of Recall}\label{sec:recall}
\input{table/recall_gen_cert}

The recall is calculated on all types of responses. Thus, we directly report the recall in \Cref{tab:recall_gen_cert}.




\subsection{Case Studies}
\label{appendix:case-study}
We present the detailed cases of different error types in \Cref{citation-guide}.

\input{table/examples}

\input{sections/6-human-evaluation}

\section{Prompts}

\subsection{Discriminative Evaluation Prompts}\label{appendix:dis_prompt}
\input{figures/discriminative_eval_prompts}

We present the zero-shot, zero-shot CoT, few-shot, and few-shot CoT prompts in \Cref{fig:dis_zero_shot_prompt,fig:dis_zero_shot_cot_prompt,fig:dis_few_shot_prompt,fig:dis_few_shot_cot_prompt}.

\subsection{Generative Evaluation Prompts}
\label{appendix:gen-prompt}
We present the few-shot, few-shot plan prompts in \Cref{fig:fewshot_cot_only_prompt,fig:fewshot_cot_hint_prompt}.

\input{figures/fewshot-cot-only-prompt}

\input{figures/fewshot-cot-hint-prompt}


%% file: sections/appx/appx-generative-eval.tex
\section{Ground-truth Reasoning Path Construction}
\label{appendix:subgraph_gr_construct} 
To obtain graph and ground truth reasoning paths for each question, we utilize SPARQL query, topic entities, answer entities provided in datasets to construct subgraphs, then find paths from topic entities to answer entities as the ground-truth reasoning paths. In particular, 
\begin{itemize}
    \item We replace \texttt{SELECT} query in each SPARQL query into a \texttt{CONSTRUCT} query to return the corresponding graph. An example \texttt{CONSTRUCT} query and its returned graph from the CWQ dataset is shown in example \ref{example:sparql_construct}.

    \item In the returned graph, we use NetworkX\footnote{\url{https://networkx.org/}} library to find the paths starting at one of the topic entities and ending at one of the answer entities provided in original dataset. As a result, we obtain a list of ground-truth reasoning paths used for both discriminative and generative evaluation.
\end{itemize}

\begin{example}
An example \texttt{CONSTRUCT} query and the corresponding returned graph.\\

    \ttfamily
    \small
    \noindent\hrule
    CONSTRUCT query
    \hrule \noindent\\
    PREFIX ns: <http://rdf.freebase.com/ns/>\\
CONSTRUCT WHERE \{ FILTER (?x != ?c)\\
FILTER (!isLiteral(?x) OR lang(?x) = ” OR langMatches(lang(?x), ’en’))\\
?c ns:location.country.administrative\_divisions ns:m.02g\_\_4 .\\
?c ns:location.country.languages\_spoken ?x .\\
\}
\hrule
Corresponding returned graph
\hrule \noindent\\
m.03gyl location.country.administrative\_divisions m.02g\_\_4 .\\
m.03gyl location.country.languages\_spoken m.02dhwl . \\
m.03gyl location.country.languages\_spoken m.064\_8sq .
\hrule
\label{example:sparql_construct}
\end{example}

\section{Generative Evaluation}\label{app:gen_eval}
The details of triple retrieval, reasoning path evaluation, and fine-grained path evaluation are shown in the \Cref{algo:retrieval,algo:eval,algo:eval_groundtruth}, respectively.

\input{algos/algo-retrieval}

\input{algos/algo-eval}

\input{algos/algo_edit_eval}

%% file: algos/algo-retrieval.tex
\begin{algorithm}[t]
\caption{Triple Retrieval Algorithm }
\label{algo:retrieval}
    \SetAlgoLined
    \DontPrintSemicolon
    \KwIn{Step $s$; Top-$K$;
       embedding model $E$;
       knowledge graph $\mathcal{G}$. 
    }
    \KwOut{
        Triple $T$ 
    }
$h_s \gets E(s)$\; 

$[T_i, \tau_i]_{i=1}^K \gets \mathcal{G}.\text{search}(h_s, k)$\; 
$\mathcal{C} = []$\;
\For{$i \gets 1$ \KwTo $K$}{
    $\epsilon_h,\epsilon_t \gets \text{Fuzzy-match}(T_i, s)$\;
    $\tilde{\tau_i} \gets \frac{\tau_i + \epsilon_h + \epsilon_t}{3}$\;
    $\mathcal{C}.\text{append}((\tilde{\tau_i},T_i))$
}
$T = \underset{T_i\in \mathcal{C}}{\arg\max}~\tilde{\tau_i}$\;
\end{algorithm}

%% file: algos/algo-eval.tex
\begin{algorithm}[t]
\small
\caption{Reasoning Path Evaluation }
\label{algo:eval}
    \SetAlgoLined
    \DontPrintSemicolon
    \KwIn{
        Reasoning path $\hat{P}$;
        threshold $\sigma$; knowledge graph $\mathcal{G}$; ground answer entity $\mathcal{A}$.
    }
    \KwOut{Validity $v$.
        }
$\text{factual\_error} \gets False$\;





$order\_error \gets  False$

$answer\_error \gets  answer_{\hat{P}}!=\mathcal{A}$

\For{$T_i \in \hat{P}$}{


    \eIf {$\tau_i < \sigma$ }{
        $\text{factual\_error} \gets True$\; 
    }
    {
        \sout{continue} \\
        \If {$head_{T_i}!= tail_{T_i-1}$}{$order\_error \gets  True$}
    }
}
\BlankLine
$coherent\_error \gets \neg factual\_error \wedge order\_error$\\
$answer\_error \gets \neg coherent\_error \wedge answer\_error$

\eIf {$factual\_error \vee coherent\_error \vee answer\_error$}{
    $v \gets False$\;
}
{

    $v \gets True$
}

\Return{$v$}\;
\end{algorithm}

%% file: algos/algo_edit_eval.tex
\begin{algorithm}[t]
\caption{Fine-grained Path Evaluation }
\label{algo:eval_groundtruth}
    \SetAlgoLined
    \DontPrintSemicolon
    \KwIn{
        Reasoning path $\hat{P}$;
        Ground-truth paths $\mathcal{P}^*$
    }
    \KwOut{Edit distance $u$}
$m\gets 0$\;
$u\gets 0$\;


\For{$P^*\in\mathcal{P}^*$}{
    
    $u' \gets \text{NeedlemanWunsch}(\hat{P}, P^*)$\; 

    \If {$u' < u$}{
    $u \gets u'$\textcolor{blue}{\Comment{Get minimum edit distance.}}\; 
    }

    
    
    
    




}

\Return{$u$}
\end{algorithm}

%% file: sections/appx/appx-experiment-setting.tex
\section{Experiment Settings}
\subsection{Datasets}
\label{appendix:dataset}
We adopt two benchmark KGQA datasets: Complex WebQuestions (CWQ)\footnote{\url{https://www.tau-nlp.sites.tau.ac.il/compwebq}}\cite{Talmor_2018} and GrailQA\footnote{\url{https://huggingface.co/datasets/grail_qa}}\cite{gu2021beyond} in this work, especially we use the test split of CWQ, and validation split of GrailQA. We only keep questions requiring more than 2-hop reasoning. The number of questions is shown in \Cref{tab:stat_data}. Both CWQ and GrailQA can be reasoned based on Freebase KGs\footnote{\url{https://github.com/microsoft/FastRDFStore}}\cite{bollacker2008freebase}. To reduce the KG size, we combine the subgraphs obtained from SPARQL queries as the final KG. The detail of subgraphs extraction is shown in \Cref{appendix:subgraph_gr_construct}.

\subsection{Large Language Models}
\label{appendix:llm}

The LLMs used in experiments are shown in \Cref{tab:llm}. We utilize available checkpoints from HuggingFace\footnote{\url{https://huggingface.co/}}.

\input{table/model_implementation}

\subsection{Evaluation Metrics}
\label{appendix:eval-metric}
\paragraph{Faithful Reasoning Score.} To enable robust and truthful reasoning, many LLMs adopt the guardrail techniques to \textit{abstain} from providing answers when they are uncertain \cite{liu2023trustworthy,luo-etal-2023-systematic}.
Also, LLMs exhibit different instruction-following capabilities, which may result in the CoT response in \textit{unstructured} format. Thus, we need to consider both the \textit{abstained} and \textit{unstructured} responses in the evaluation. We define the precision and recall as, 
\begin{align}
    &\resizebox{.6\columnwidth}{!}{
    $\text{\small Precision} =\frac{\#\text{correct}}{\#\text{correct}+\#\text{incorrect}},$}\label{eq:precision}\\
    &\resizebox{\columnwidth}{!}{
    $\text{\small Recall} = \frac{\#\text{correct}}{\#\text{correct}+\#\text{incorrect}+\#\text{abstained}+\#\text{unstructured}}.$ \label{eq:recall}
    }
\end{align}
The F1 score is then the harmonic average of the precision and recall. 
The details of abstained and unstructured response detection are  in \Cref{appendix:abstained,appendix:unstructured}.

\paragraph{Answer Accuracy} The answers generated by LLM can have a different length than the ground-truth, so we utilize fuzzy matching  to check the correctness of answer. It checks whether the generated  answer appears in the ground-truth answers, and vice versa. 

\paragraph{Edit Distance} 
We further look into the fine-grained evaluation of the reasoning path by calculating the edit distance \cite{needleman1970general}: The minimum number of edits (add/remove the reasoning steps) required to convert the CoT path to the ground-truth path.

\subsection{Abstained Answer Detection}
\label{appendix:abstained}
We detect abstained answers through following specific keywords in LLMs' responses:

\begin{minipage}{0.9\columnwidth}
  \centering
  \begin{tcolorbox}[title=List of Abstention keywords, fonttitle=\bfseries\small]
      not have knowledge\\
      more information\\
      need more\\
      unknown\\
      cannot\\
      sorry \\
      impossible\\not possible\\unable\\unclear
  \end{tcolorbox}
  \vspace{1mm}
\end{minipage}

\subsection{Unstructured Answer Detection}
\label{appendix:unstructured}

As the instruction shown in prompts \ref{fig:fewshot_cot_only_prompt} and \ref{fig:fewshot_cot_hint_prompt}, a structured CoT should follow the following format: 

\begin{minipage}{0.9\columnwidth}
  \centering
  \begin{tcolorbox}[title=A structured answer, fonttitle=\bfseries\small]
        1. <step1>\\
        2. <step2>\\
        ...\\
        (T|t)he answer( to the question)? is\:? \(?(.*?)\)?.?
  \end{tcolorbox}
  \vspace{1mm}
\end{minipage}
Therefore, the CoT responses that do not match the pattern are identified as unstructured.


%% file: table/model_implementation.tex
\begin{table}
\centering
\resizebox{\columnwidth}{!}{%
\begin{tabular}{|l|r|}
\hline
\textbf{LLM}        & \textbf{Model Implementation}    \\
\hline
Mistral 7B & mistralai/Mistral-7B-Instruct-v0.1   \\
QWen 7B    & Qwen/Qwen-7B-Chat              \\
QWen 14B   & Qwen/Qwen-14B-Chat             \\
Vicuna 33B   & lmsys/vicuna-33b-v1.3             \\
LlaMA2 70B   & meta-llama/Llama-2-70b-chat            \\
ChatGPT   & GPT-3.5-turbo             \\
\hline
\end{tabular}
}
\caption{Details of used LLMs.}
\label{tab:llm}
\end{table}

%% file: table/detailed_discriminative_evaluation.tex
\begin{table*}[]
\resizebox{\textwidth}{!}{%
\begin{tabular}{@{}l|c|cccc|cccc@{}}
\toprule
\multirow{2}{*}{LLMs} & \multirow{2}{*}{Size} & \multicolumn{4}{c|}{Valid Reasoning Path}           & \multicolumn{4}{c}{Factual Errors Reasoning Path}   \\ \cmidrule(l){3-10} 
                      &                       & Zero-shot & Zero-shot-Cot & Few-shot & Few-shot CoT & Zero-shot & Zero-shot-Cot & Few-shot & Few-shot CoT \\ \midrule
 \texttt{Mistral}               & 7B                    & 79.01   & 72.32       & 98.78  & 74.54      & 98.68   & 99.46       & 74.72  & 90.67      \\
 \texttt{Qwen}                  & 7B                    & 94.92   & 91.84       & 85.05  & 58.60      & 94.71   & 94.45       & 95.25  & 92.10      \\
 \texttt{Qwen}                  & 14B                   & 62.41   & 61.97       & 75.74  & 21.69      & 100.00  & 99.96       & 99.78  & 99.31      \\
 \texttt{Vicuna}                & 33B                   & 83.54   & 79.32       & 91.80  & 19.02      & 99.37   & 99.77       & 98.96  & 98.40      \\
 \texttt{LLaMA2}                & 70B                   & 18.79   & 27.84       & 92.50  & 57.52      & 99.54   & 99.50       & 59.44  & 72.48      \\
 \texttt{ChatGPT}               & 175B                  & 70.27   & 74.57       & 86.56  & 71.45      & 99.72   & 99.72       & 99.70  & 95.86      \\ \midrule
\multirow{2}{*}{LLMs} & \multirow{2}{*}{Size} & \multicolumn{4}{c|}{Incoherent Reasoning Path}      & \multicolumn{4}{c}{Misguided Reasoning Path}        \\ \cmidrule(l){3-10} 
                      &                       & Zero-shot & Zero-shot-Cot & Few-shot & Few-shot CoT & Zero-shot & Zero-shot-Cot & Few-shot & Few-shot CoT \\ \midrule
 \texttt{Mistral}               & 7B                    & 90.11   & 92.10       & 39.92  & 53.08      & 82.55   & 95.62       & 14.20  & 61.61      \\
 \texttt{Qwen}                  & 7B                    & 61.37   & 61.82       & 73.06  & 66.37      & 48.02   & 56.40       & 65.18  & 75.85      \\
 \texttt{Qwen}                  & 14B                   & 93.56   & 94.91       & 88.56  & 85.49      & 98.38   & 98.61       & 91.14  & 97.00      \\
 \texttt{Vicuna}                & 33B                   & 90.21   & 93.62       & 65.65  & 56.56      & 98.04   & 98.82       & 83.21  & 94.23      \\
 \texttt{LLaMA2}                & 70B                   & 95.92   & 97.43       & 44.41  & 18.27      & 97.58   & 98.05       & 31.60  & 42.78      \\
 \texttt{ChatGPT}               & 175B                  & 93.36   & 90.57       & 76.76  & 59.01      & 96.07   & 95.82       & 85.34  & 94.28      \\ \bottomrule
\end{tabular}%
}
\caption{Discriminative evaluation results of different LLMs on CWQ dataset. We use the binary accuracy as metric.}\label{dis_result}
\end{table*}

%% file: figures/prec_cot_only.tex
\begin{figure*}
   \centering
   \resizebox{\textwidth}{!}{%
    \begin{tabular}{cc}
        \includegraphics[width=8cm]{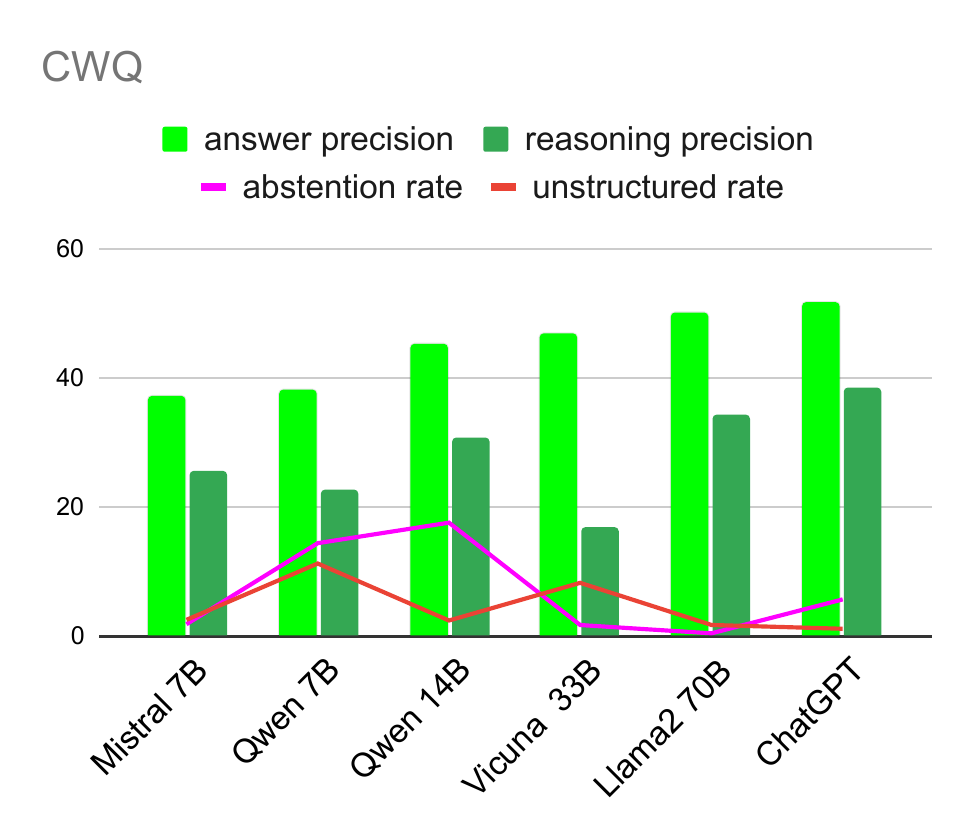}
        &
        \includegraphics[width=8cm]{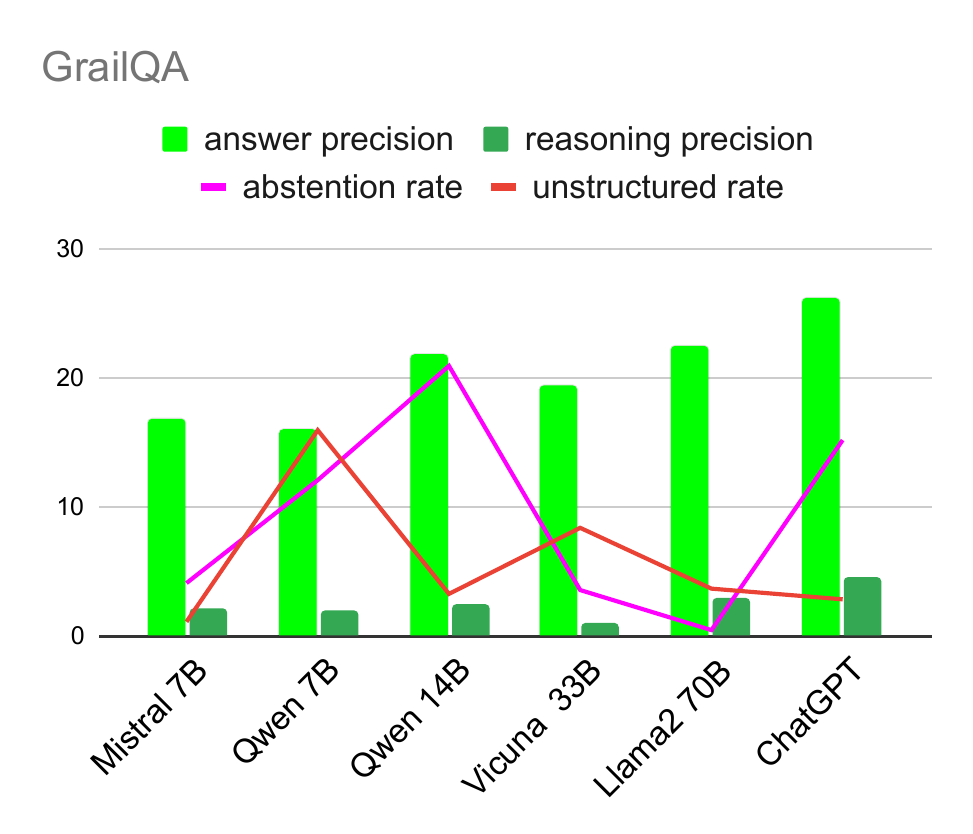}
    \end{tabular}
    }
    \caption{The precision of LLMs using few-shot CoT prompt.}
    \label{fig:prec_cot_only}
\end{figure*}

%% file: figures/prec_cot_hint.tex
\begin{figure*}
   \centering
   \resizebox{\textwidth}{!}{%
    \begin{tabular}{cc}
        \includegraphics[width=8cm]{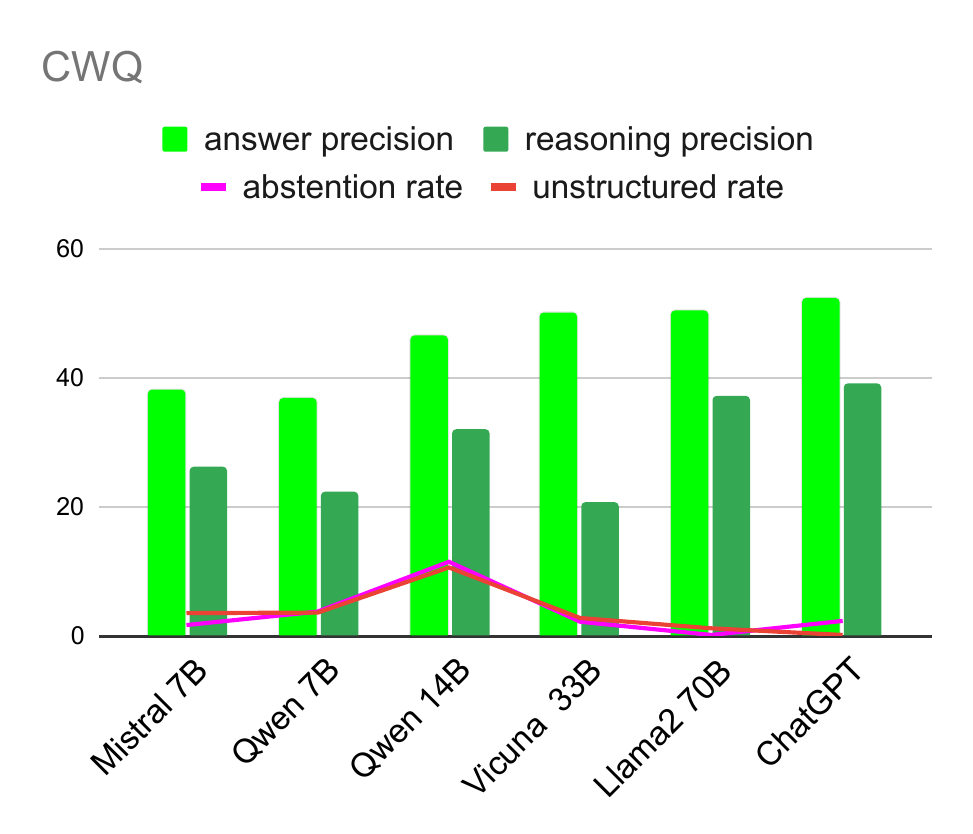}
        &
        \includegraphics[width=8cm]{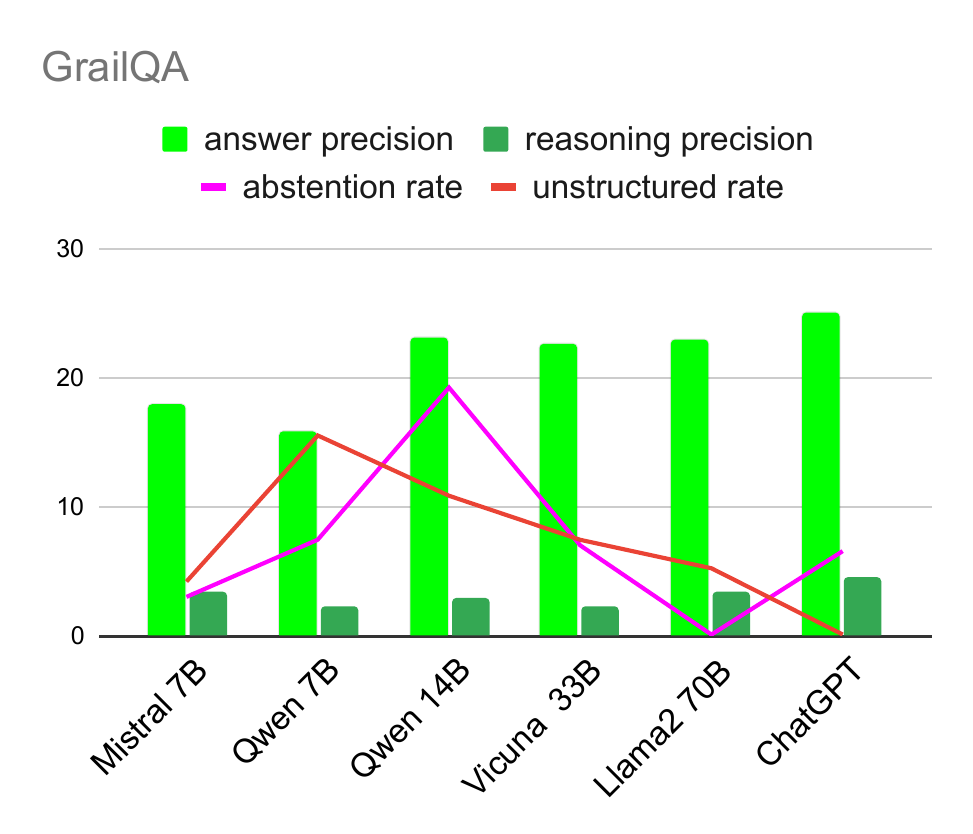}
    \end{tabular}
    }
    \caption{The precision of LLMs using few-shot CoT - Plan prompt.}
    \label{fig:prec_hint}
\end{figure*}

%% file: figures/prec_cot_sc.tex
\begin{figure*}
   \centering
   \resizebox{\textwidth}{!}{%
    \begin{tabular}{cc}
        \includegraphics[width=8cm]{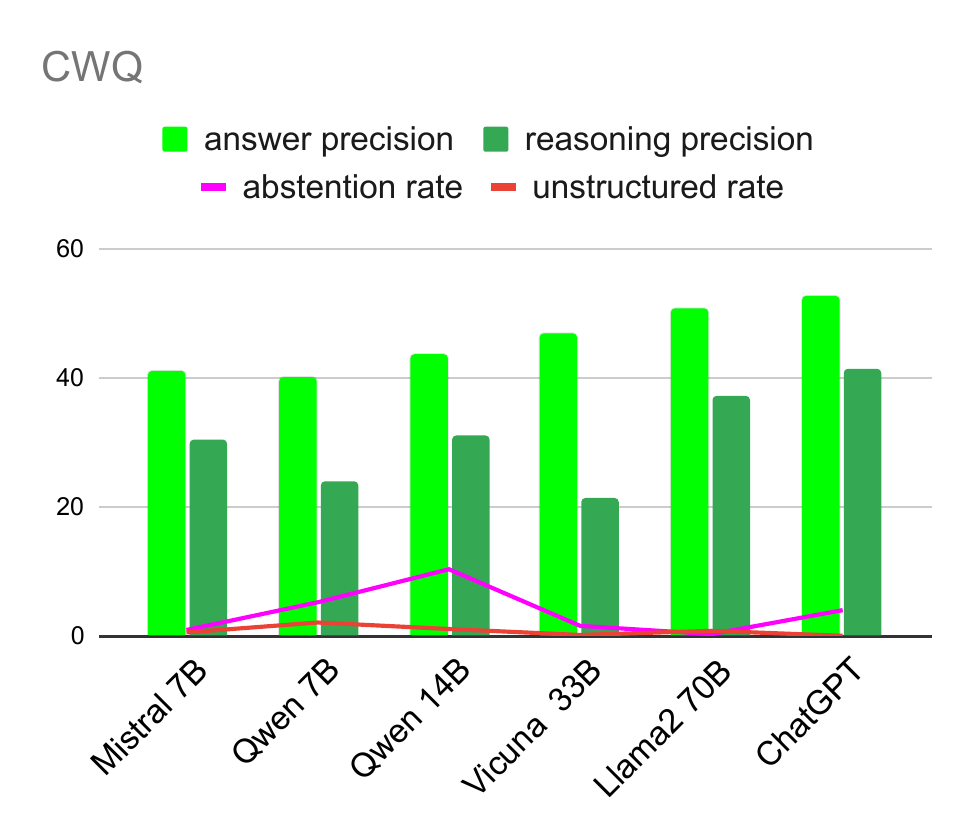}
        &
        \includegraphics[width=8cm]{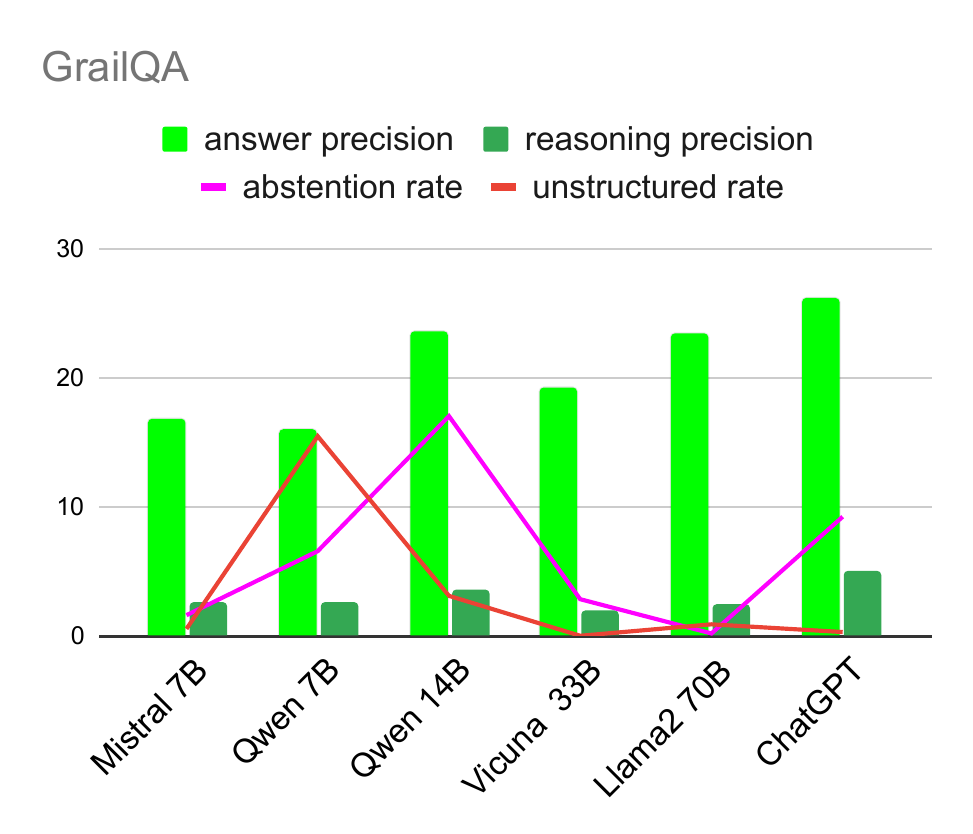}
    \end{tabular}
    }
    \caption{The precision of LLMs using few-shot CoT - SC prompt.}
    \label{fig:prec_cot_sc}
\end{figure*}

%% file: table/recall_gen_cert.tex
\begin{table}
\centering
\small
\resizebox{\columnwidth}{!}{%
\begin{tabular}{@{}c|c|cc|cc@{}}  
\toprule
\multirow{2}{*}{LLMs} & \multirow{2}{*}{Size} & \multicolumn{2}{c|}{\textbf{CWQ}} & \multicolumn{2}{c}{\textbf{GrailQA}}\\
\cmidrule(l){3-6} 
 &  & \multicolumn{1}{c}{Answer} & Reasoning & 
\multicolumn{1}{c}{Answer} & \multicolumn{1}{c}{Reasoning}\\
\midrule 
& & \multicolumn{4}{c}{Fewshot CoT} \\  
\midrule 
 Mistral & 7B & 35.67 & 24.64 & 15.93 & 2.07\\ 
 Qwen & 7B & 28.36 & 16.9  & 11.48 & 1.4\\
 Qwen  & 14B &  36.37 & 24.65 & 16.55 & 1.87\\
 Vicuna & 33B &  42.3 & 15.13 & 17.17 & 0.89\\
LLaMA2 & 70B & 49.29 & 33.63 & 21.6 & 2.82 \\
ChatGPT & 175B & 48.17 & 35.88 & 21.55 & 3.8\\  
\midrule
& & \multicolumn{4}{c}{Fewshot CoT - Plan} \\ 
\midrule
 Mistral & 7B & 36.18 & 25.02 &  16.67 & 3.24\\
 Qwen  & 7B & 34.13 & 20.78 & 12.16 & 1.82\\
 Qwen  & 14B & 36.35 & 24.88 & 16.14 & 2.11\\
 Vicuna & 33B & 47.6 & 19.74 & 19.33 & 1.94\\
LLaMA2  & 70B & 49.94 & 36.84 & 21.76 & 3.2\\ 
ChatGPT & 175B & 51.11 & 38.13 & 23.4 & 4.18\\ 
 \midrule
& & \multicolumn{4}{c}{Fewshot CoT - SC} \\ 
\midrule
 Mistral  & 7B &  40.57 & 30.17 &  16.52 & 2.57\\
 Qwen  & 7B & 37.2 & 22.37 & 12.45 & 2.07\\
 Qwen  & 14B & 38.68 & 27.6 & 18.88 & 2.91\\
 Vicuna & 33B & 46.29 & 21.14 & 18.65 & 1.86\\
LLaMA2 & 70B & 50.16 & 36.81 & 21.76 & 3.2\\ 
ChatGPT & 175B & 50.71 & 39.92 & 23.78 & 4.63\\ 
 \bottomrule
\end{tabular}
}
\caption{
The recall of LLMs on CWQ and GrailQA datasets.
}
\label{tab:recall_gen_cert}
\end{table}

%% file: table/examples.tex
\begin{table*}
\centering
\resizebox{\textwidth}{!}{%
\begin{tabular}{ll}
\hline
Correct Answer \& Correct Reasoning & 
\begin{tabular}[c]{@{}l@{}}Question: \\ People from the country that contains Corfu speak what language?\\ \\ Groundtruth: \\ {[}{[}"Corfu","location.country.administrative\_divisions","Greece"{]},  \\ {[}"Greece","location.country.languages\_spoken","Greek Language"{]}{]}\\ \\ LLM CoT Answer: \\ 1. Corfu is an island located in Greece.\\ 2. Greece is a country that contains Corfu.\\ 3. The primary language spoken in Greece is Greek.\\ So the answer is (Greek).\\\\ Reasoning path from our framework:\\ 1. {[}{[}"Corfu","location.country.administrative\_divisions","Greece"{]}
\\ 2. "Greece","location.country.languages\_spoken","Greek Language"{]}{]}
\end{tabular}
\\ \hline
Correct Answer \& Incorrect Reasoning & 
\begin{tabular}[c]{@{}l@{}}Question: \\ What is the currency in the country where the Brazilian cruzado is also used?\\ \\ Groundtruth:\\ {[}{[}"Brazilian cruzado","location.country.currency\_formerly\_used","Brazil"{]}, \\ {[}"Brazil","location.country.currency\_used","Brazilian real"{]}{]}\\ \\ LLM CoT Answer: \\ \textcolor{red}{1. The Brazilian cruzado was the currency of Brazil from 1986 to 1989.}\\ \textcolor{red}{2. Brazil is the country where the Brazilian cruzado was used.}\\ So the answer is (\textcolor{green}{Brazilian real}).\\ \\ Reasoning path from our framework:\\ 1. {[}"Brazilian cruzado","location.country.currency\_formerly\_used","Brazil"{]}
\\ 2. {[}"Brazilian cruzado","location.country.currency\_formerly\_used","Brazil"{]}
\end{tabular}
\\ \hline
Incorrect Answer \& Correct Reasoning & 
\begin{tabular}[c]{@{}l@{}}Question: \\ Where was the main artist featured in the Rihanna: Live in Concert Tour raised?\\ \\ Groundtruth: \\ {[}{[}"Rihanna: Live in Concert Tour","music.artist.concert\_tours","Rihanna"{]}, \\ {[}"Rihanna","people.person.place\_of\_birth","\textcolor{green}{Saint Michael Parish}"{]}{]}\\ \\ LLM CoT Answer:\\ 1. The main artist featured in the Rihanna: Live in Concert Tour is Rihanna.\\ 2. Rihanna was raised in Barbados.\\ So the answer is (\textcolor{red}{Barbados}).\\ \\ Reasoning path from our framework:\\ 1. {[}"Rihanna: Live in Concert Tour","music.artist.concert\_tours","Rihanna"{]}
\\ 2. {[}"Rihanna","people.person.nationality","Barbados"{]}
\end{tabular}
\\ \hline
Incorrect Answer \& Incorrect Reasoning & 
\begin{tabular}[c]{@{}l@{}}Question: \\ Where did the author who published "The Story of My Life" die?\\ \\ Groundtruth:\\ {[}{[}"The story of my life","book.author.book\_editions\_published","Helen Keller"{]}, \\ {[}"Helen Keller","people.deceased\_person.place\_of\_death","\textcolor{green}{Easton}"{]}{]}\\ \\ LLM CoT Answer: \\ 1. "The Story of My Life" is a book written by Helen Keller.\\ \textcolor{red}{2. Helen Keller died in Connecticut, USA.}\\ So the answer is (\textcolor{red}{Connecticut, USA}).\\ \\ Reasoning path from our framework:\\ 1. {[}"The story of my life","book.author.book\_editions\_published","Helen Keller"{]}, 
\\ 2. {[}"Helen Keller","people.deceased\_person.place\_of\_death","Easton"{]},\textcolor{red}{p=0.57} 
\end{tabular}
\\
\hline
\end{tabular}
}
\caption{\label{citation-guide}
Examples of answer and reasoning accuracy of generative evaluation framework.
}
\end{table*}

%% file: sections/6-human-evaluation.tex
\section{Human Evaluation}
\label{sec:human_eval}
This section aims to be two-fold. Firstly, it provides a sanity check of the proposed prompt in instructing LLMs to generate CoT in a structured format. Secondly, we aim to construct a small test set to evaluate the capability of the proposed framework in detecting the faithful CoT.
\subsection{Annotation}
Firstly, a sample of 100 questions is chosen randomly from the CWQ dataset along with the responses generated by ChatGPT using proposed prompts. Each question in the sample is matched with reference answers, which include the final answer provided in the datasets, as well as all the ground-truth reasoning paths extracted during the preprocessing phase outlined in appendix \ref{appendix:subgraph_gr_construct}. This data is supplied to annotators for classification purposes.

The annotators are two PhD students who are familiar with the related works and have experience working with knowledge graphs. To prepare them for the task, we provide them the definition of faithful reasoning and error types. Given sample cases that consist of the question, answer, the ground-truth reasoning paths (the desired reasoning steps), and the model's initial CoT response, two human experts performed a trial run to align their understanding and coding criteria for faithful CoT responses by answering two following questions:
\begin{itemize}
    \item \texttt{Q1: If LLM response is an incorrect final answer?} True labeling if LLM final answer does not hit the reference answers; False otherwise.
    \item \texttt{Q2: If LLM CoT is an incorrect reasoning?} True labeling if there exists any step its relation, subject/object is in ground-truth reasoning paths but there is no link between them or the object/subject/relation is not in the ground-truth reasoning paths; False otherwise.
\end{itemize}

The human annotation dataset acquired ultimately comprises 100 samples, with each sample consisting of the question, ground truth paths, LLM answers, a column indicating the annotated incorrect answer (1 for True labeling of Q1, 0 otherwise), and another column indicating the annotated incorrect reasoning (1 for True labeling of Q2, 0 otherwise). 

\subsection{Framework Evaluation} 
Given the human annotation data provided above, our generative evaluation framework is assessed to ensure its quality prior to its application in all experiments. Initially, the responses from the LLM are inputted into the generative evaluation mode to obtain predictions for Q1 (predicted incorrect answer) and Q2 (predicted incorrect reasoning). Consequently, these two outcomes are compared with the annotated dataset to assess the precision, recall, and f1\_score for the final score. The metrics are shown in table \ref{tab:eval_framework}.

\color{black}

%% file: figures/discriminative_eval_prompts.tex
\begin{figure*}  
  \begin{minipage}[l]{1.\textwidth}
  \centering
    \begin{tcolorbox}[title=Zero-shot Discriminative Evaluation Prompt, fonttitle=\bfseries]
      A reasoning path is a sequence of triples that can be used to derive the answer of given question. A valid reasoning path should follow these rules:
      \vspace{5pt}
      \\
1. No factual errors: Each triple in the reasoning path should adhere to real-world factual knowledge.\\
2. Coherence: The tail entity of the previous triple should be the head entity of the next triple.\\
3. Correctness: The reasoning path should lead to the correct answer at the last tail entity.
\vspace{5pt}
      \\
Given this reasoning path, do you think this is a valid path to derive the answer of given question? If yes please answer "YES", otherwise please answer "NO".
      \vspace{5pt}
      \\
      Question:\\
      \texttt{<Question>}
      \vspace{5pt}
      \\
      Answer:\\
     \texttt{<Answer>}
      \vspace{5pt}
      \\
      Reasoning path:\\
      \texttt{<Reasoning Path>}
  \end{tcolorbox}
  \vspace{1mm}
\end{minipage}
\caption{The zero-shot prompt used for discriminative evaluation}
\label{fig:dis_zero_shot_prompt}
\end{figure*}

\begin{figure*}  
  \begin{minipage}[l]{1.\textwidth}
  \centering
    \begin{tcolorbox}[title=Zero-shot CoT Discriminative Evaluation Prompt, fonttitle=\bfseries]
    A reasoning path is a sequence of triples that can be used to derive the answer of given question. A valid reasoning path should follow these rules:
      \vspace{5pt}
      \\
1. No factual errors: Each triple in the reasoning path should adhere to real-world factual knowledge.\\
2. Coherence: The tail entity of the previous triple should be the head entity of the next triple.\\
3. Correctness: The reasoning path should lead to the correct answer at the last tail entity.
\vspace{5pt}
      \\
      Given this reasoning path, do you think this is a valid path to answer the question? If yes please answer "YES", otherwise please answer "NO". Let's think it step by step.
      \vspace{5pt}
      \\
      Question:\\
      \texttt{<Question>}
      \vspace{5pt}
      \\
        Answer:\\
     \texttt{<Answer>}
           \vspace{5pt}
      \\
      Reasoning path:\\
      \texttt{<Reasoning Path>}
  \end{tcolorbox}
  \vspace{1mm}
\end{minipage}
\caption{The zero-shot CoT prompt used for discriminative evaluation.}
\label{fig:dis_zero_shot_cot_prompt}
\end{figure*}

\begin{figure*}  
  \begin{minipage}[l]{1.\textwidth}
  \centering
    \begin{tcolorbox}[title=Few-shot Discriminative Evaluation Prompt, fonttitle=\bfseries]
    \small
    A reasoning path is a sequence of triples that can be used to derive the answer of given question. A valid reasoning path should follow these rules:
      \vspace{5pt}
      \\
1. No factual errors: Each triple in the reasoning path should adhere to real-world factual knowledge.\\
2. Coherence: The tail entity of the previous triple should be the head entity of the next triple.\\
3. Correctness: The reasoning path should lead to the correct answer at the last tail entity.
\vspace{5pt}
      \\
      Given this reasoning path, do you think this is a valid path to answer the question? If yes please answer "YES", otherwise please answer "NO". Here are some examples:
    \vspace{5pt}
    \\
    \#\# Input:
    \\
    Question: \\
    What type of government is used in the country with Northern District?
    \vspace{5pt}
    \\
    Answer:\\
    Parliamentary system
    \vspace{5pt}
    \\
    Reasoning Paths: \\
    Step 1: Northern District -> location.administrative\_division.first\_level\_division\_of -> Israel \\
    Step 2: Israel -> government.form\_of\_government.countries -> Parliamentary system
    \vspace{5pt}
    \\
    \#\# Output:
    \\
    YES
    \vspace{5pt}
    \\
    \#\# Input:\\
    Question: \\
    Where is the home stadium of the team who won the 1946 World Series championship?
    \vspace{5pt}
    \\
    Answer:\\
    Busch Stadium
    \vspace{5pt}
    \\
    Reasoning Paths: \\
    Step 1: 1946 World Series -> sports.sports\_team.championships -> St. Louis Cardinals \\
    Step 2: St. Louis Cardinals -> sports.sports\_team.arena\_stadium -> Roger Dean Stadium
    \vspace{5pt}
    \\
    \#\# Output:\\
    NO
    \vspace{5pt}
    \\
    \#\# Input: \\
    Question:\\
    In which American Southern City did the ""Downs"" composer die?
    \vspace{5pt}
    \\
    Answer:\\
    New Orleans
    \vspace{5pt}
    \\
    Reasoning Paths:\\
    Step 1: Alex Chilton -> people.deceased\_person.place\_of\_death -> New Orleans \\
    Step 2: Downs -> music.composer.compositions -> Alex Chilton
    \vspace{5pt}
    \\
    \#\# Output:\\
    NO
    \vspace{5pt}
    \\
    \#\# Input:\\
    Question:\\
    Where was the main artist featured in the Rihanna: Live in Concert Tour raised?
    \vspace{5pt}
    \\
    Answer: \\
    Saint Michael Parish
    \vspace{5pt}
    \\
    Reasoning Paths:\\
    Step 1: Rihanna: Live in Concert Tour -> music.concert\_tour.artist -> Rihanna\\
    Step 2: Rihanna -> music.track\_contribution.role -> Vocals
    \vspace{5pt}
    \\
    \#\# Output:
    NO
    \vspace{5pt}
    \\
    \#\# Input:\\
    Question:\\
    \texttt{<Question>}
    \vspace{5pt}
    \\
        Answer:\\
     \texttt{<Answer>}
           \vspace{5pt}
      \\
    Reasoning path:\\
    \texttt{<Reasoning Path>}
    \vspace{5pt}
    \\
    \#\# Output:
  \end{tcolorbox}
  \vspace{1mm}
\end{minipage}
\caption{The few-shot prompt used for discriminative evaluation.}
\label{fig:dis_few_shot_prompt}
\end{figure*}

\begin{figure*}  
  \begin{minipage}[l]{1.\textwidth}
  \scriptsize
    \begin{tcolorbox}[title=Few-shot CoT Discriminative Evaluation Prompt, fonttitle=\bfseries]
    \tiny
    A reasoning path is a sequence of triples that can be used to derive the answer of given question. A valid reasoning path should follow these rules:
      \vspace{5pt}
      \\
1. No factual errors: Each triple in the reasoning path should adhere to real-world factual knowledge.\\
2. Coherence: The tail entity of the previous triple should be the head entity of the next triple.\\
3. Correctness: The reasoning path should lead to the correct answer at the last tail entity.
\vspace{5pt}
      \\
      Given this reasoning path, do you think this is a valid path to answer the question? If yes please answer "YES", otherwise please answer "NO". Here are some examples:
    \vspace{5pt}
    \\
    \#\# Input:
    \\
    Question: \\
    What type of government is used in the country with Northern District?
    \vspace{5pt}
    \\
    Answer:\\
    Parliamentary system
    \vspace{5pt}
    \\
    Reasoning Paths: \\
    Step 1: Northern District -> location.administrative\_division.first\_level\_division\_of -> Israel \\
    Step 2: Israel -> government.form\_of\_government.countries -> Parliamentary system
    \vspace{5pt}
    \\
    \#\# Output:
    \\
    This reasoning path indicates that:\\
1. "Northern District" is a location within some country.\\
2. The reasoning path mentions "Northern District -> location.administrative\_division.first\_level\_division\_of -> Israel," indicating that the Northern District is part of Israel.\\
3. It further states "Israel -> government.form\_of\_government.countries," suggesting that Israel's form of government is being discussed.\\
4. The last part of the reasoning path indicates that Israel has a "Parliamentary system."
\\
Based on the provided reasoning paths, it can be concluded that the type of government used in the country with the Northern District (Israel) is a Parliamentary system. Therefore, the path adheres to the definition of valid reasoning path. The answer is "YES"
    \vspace{5pt}
    \\
    \#\# Input:\\
    Question: \\
    Where is the home stadium of the team who won the 1946 World Series championship?
    \vspace{5pt}
    \\
    Answer:\\
    Busch Stadium
    \vspace{5pt}
    \\
    Reasoning Paths: \\
    Step 1: 1946 World Series -> sports.sports\_team.championships -> St. Louis Cardinals \\
    Step 2: St. Louis Cardinals -> sports.sports\_team.arena\_stadium -> Roger Dean Stadium
    \vspace{5pt}
    \\
    \#\# Output:\\
    This reasoning path indicates that:\\
1. St. Louis Cardinals as the team that won the 1946 World Series\\
2. Roger Dean Stadium is the stadium associated with the St. Louis Cardinals.
\\
However, the home stadium of St. Louis Cardinals is Busch Stadium, not Roger Dean Stadium. Therefore, this path contains factual errors. The answer is "NO".
    \vspace{5pt}
    \\
    \#\# Input: \\
    Question:\\
    In which American Southern City did the ""Downs"" composer die?
    \vspace{5pt}
    \\
    Answer:\\
    New Orleans
    \vspace{5pt}
    \\
    Reasoning Paths:\\
    Step 1: Alex Chilton -> people.deceased\_person.place\_of\_death -> New Orleans \\
    Step 2: Downs -> music.composer.compositions -> Alex Chilton
    \vspace{5pt}
    \\
    \#\# Output:\\
    This reasoning path indicates that:\\
1. Alex Chilton was dead in New Orleans\\
2. The composition of Downs is Alex Chilton\\

Even through the first step of reasoning path leads to the correct answer, which is New Orleans, the reasoning path is not coherent. To answer the question, we need to first the composition of Downs which is Alex Chilton, then we find the death place of Alex Chilton which is New Orleans. Therefore, the answer is "NO".
    \vspace{5pt}
    \\
    \#\# Input:\\
    Question:\\
    Where was the main artist featured in the Rihanna: Live in Concert Tour raised?
    \vspace{5pt}
    \\
    Answer: \\
    Saint Michael Parish
    \vspace{5pt}
    \\
    Reasoning Paths:\\
    Step 1: Rihanna: Live in Concert Tour -> music.concert\_tour.artist -> Rihanna\\
    Step 2: Rihanna -> music.track\_contribution.role -> Vocals
    \vspace{5pt}
    \\
    \#\# Output:
    This reasoning path indicates that:\\
1. The artist of Rihanna: Live in Concert Tour is Rihanna\\
2. Rihanna is a vocal artist\\

Even through there are no factual erros and the reasoning path is coherent, the reasoning path does not lead to the correct answer. The question asks for the birth place of the main artist, not the role of the artist. Therefore, the answer is "NO".
    \vspace{5pt}
    \\
    \#\# Input:\\
    Question:\\
    \texttt{<Question>}
    \vspace{5pt}
    \\
        Answer:\\
     \texttt{<Answer>}
           \vspace{5pt}
      \\
    Reasoning path:\\
    \texttt{<Reasoning Path>}
    \vspace{5pt}
    \\
    \#\# Output:
  \end{tcolorbox}
  \vspace{1mm}
\end{minipage}
\caption{The few-shot CoT prompt used for discriminative evaluation.}
\label{fig:dis_few_shot_cot_prompt}
\end{figure*}

%% file: figures/fewshot-cot-only-prompt.tex
\begin{figure*}
\begin{minipage}[l]{1.05\textwidth}
  \centering
  
  \begin{tcolorbox}[title=Few-shot CoT prompt for Generative Evaluation, fonttitle=\bfseries]
  1. <step1>\\
2. <step2>\\
...\\
So the answer is (<answer>).\\
Make sure that the answer uses the above format and answers the question step by step.\\
\\
Q: when Lou Seal is the mascot for the team that last won the World Series?\\
A: Let's work this out in a step by step way to be sure we have the right answer.\\
1. Lou Seal is the mascot for the San Francisco Giants.\\
2. The San Francisco Giants are associated with the sports championship event, the 2014 World Series.\\
So the answer is (2014 World Series).\\
\\
Q: What nation has an army or more than 713480 people and borders the country of Bolivia?\\
A: Let's work this out in a step by step way to be sure we have the right answer.\\
1. Bolivia is a landlocked country located in South America.\\
2. Bolivia shares its borders with several countries, including Argentina, Brazil, Chile, Paraguay, and Peru.\\
So the answer is (Brazil).\\
\\
Q: What movie was displayed at the 2012 Refugee Film Festival and had Angelia Jolie directing it?\\
A: Let's work this out in a step by step way to be sure we have the right answer.\\
1. Angelia Jolie whose first major film as a director which named “In the Land of Blood and Honey”.\\
2. “In the Land of Blood and Honey” was shown at the 2012 Refugee Film Festival.\\
So the answer is (In the Land of Blood and Honey).\\
\\
Q: How many Mary Mary sisters?\\
A: Let's work this out in a step by step way to be sure we have the right answer.\\
1. Mary Mary is a group which has a member named Tina Campbell\\
2. Mary Mary is a group which has a member named Erica Campbell\\
So the answer is (Erica Campbell, Tina Campbell).\\
\\
Q: Which languages are used in the location that the breed Egyptian Mau started in?\\
A: Let's work this out in a step by step way to be sure we have the right answer.\\
1. The Egyptian Mau is a breed of domestic cat that is believed to have originated in Egypt.\\
2. In Egypt, the primary language spoken is Arabic, besides Domari or Nobiin.\\
So the answer is (Arabic, Domari, Nobiin).\\
\\
Q: \texttt{\{Question\}}\\
A: Let's work this out in a step by step way to be sure we have the right answer.\\
  \end{tcolorbox}
  \vspace{1mm}
\end{minipage}
\caption{The few-shot CoT prompt used for generative evaluation.}
\label{fig:fewshot_cot_only_prompt}
\end{figure*}

%% file: figures/fewshot-cot-hint-prompt.tex
\begin{figure*}  
  \begin{minipage}[l]{1.0\textwidth}
  \centering
  \begin{tcolorbox}[title=Few-shot CoT - Plan prompt used for Generative Evaluation, fonttitle=\bfseries]
  Relation path is a sequence relation that describes each step of the reasoning process. \\You first give a relation path as a HINT, then reason the answer step-by-step based on it.\\
HINT:\\
1. <step1>\\
2. <step2>\\
...\\
So the answer is (<answer>).\\
Make sure that the answer uses the above format and answers the question step by step.\\
\\
Q: when Lou Seal is the mascot for the team that last won the World Series?\\
A: Let's work this out in a step by step way to be sure we have the right answer.\\
HINT: sports.sports\_team.team\_mascot -> sports.sports\_team.championships\\
1. Lou Seal is the mascot for the San Francisco Giants.\\
2. The San Francisco Giants are associated with the sports championship event, the 2014 World Series.\\
So the answer is (2014 World Series).\\
\\
Q: What nation has an army or more than 713480 people and borders the country of Bolivia?\\
A: Let's work this out in a step by step way to be sure we have the right answer.\\
HINT: geography.river.basin\_countries -> location.location.partially\_contains\\
1. Bolivia is a landlocked country located in South America.\\
2. Bolivia shares its borders with several countries, including Argentina, Brazil, Chile, Paraguay, and Peru.\\
So the answer is (Brazil).\\
\\
Q: What movie was displayed at the 2012 Refugee Film Festival and had Angelia Jolie directing it?\\
A: Let's work this out in a step by step way to be sure we have the right answer.\\
HINT: film.director.film -> film.film\_regional\_release\_date.film\_regional\_debut\_venue\\
1. Angelia Jolie whose first major film as a director which named “In the Land of Blood and Honey”.\\
2. “In the Land of Blood and Honey” was shown at the 2012 Refugee Film Festival.\\
So the answer is (In the Land of Blood and Honey).\\
\\
Q: How many Mary Mary sisters?\\
A: Let's work this out in a step by step way to be sure we have the right answer.\\
HINT: music.group\_membership.member -> music.group\_membership.member\\
1. Mary Mary is a group which has a member named Tina Campbell\\
2. Mary Mary is a group which has a member named Erica Campbell\\
So the answer is (Erica Campbell, Tina Campbell).\\
\\
Q: Which languages are used in the location that the breed Egyptian Mau started in?\\
A: Let's work this out in a step by step way to be sure we have the right answer.\\
HINT: biology.breed\_origin.breeds\_originating\_here -> location.country.languages\_spoken\\
1. The Egyptian Mau is a breed of domestic cat that is believed to have originated in Egypt.\\
2. In Egypt, the primary language spoken is Arabic, besides Domari or Nobiin.\\
So the answer is (Arabic, Domari, Nobiin).\\
\\
Q: \texttt{\{Question\}}\\
A: Let's work this out in a step by step way to be sure we have the right answer.\\
  \end{tcolorbox}
  \vspace{1mm}
\end{minipage}
\caption{The few-shot CoT - Plan prompt used for generative evaluation.}
\label{fig:fewshot_cot_hint_prompt}
\end{figure*}